\documentclass[10pt,journal,compsoc]{IEEEtran}

\usepackage{epsfig}
\usepackage{graphicx}
\usepackage{amsmath}
\usepackage{amssymb}
\usepackage{mathtools}
\usepackage{bbm}
\usepackage{makecell, verbatim, sidecap}

\usepackage[nocompress]{cite}

\usepackage{balance}

\usepackage[breaklinks=true,bookmarks=false]{hyperref}

\usepackage{times}
\usepackage{epsfig}
\usepackage{graphicx}
\usepackage{amsmath}
\usepackage{amssymb}
\usepackage{bm}
\usepackage{algorithm}
\usepackage{algorithmic}
\usepackage{multirow}
\usepackage{eqparbox}
\usepackage{color}
\usepackage{float}
\usepackage{makecell}

\usepackage{color}
\usepackage[dvipsnames]{xcolor}

\renewcommand{\texttt}[1]{${{\tt #1}}$}

\usepackage{xspace}

\renewcommand{\paragraph}{\textbf}

\usepackage{graphicx}

\usepackage{graphicx}
\usepackage{amsmath}
\usepackage{amssymb}
\usepackage{booktabs}
\usepackage{bm}
\usepackage{color}
\usepackage{colortbl}
\definecolor{gray}{rgb}{0.9,0.9,0.9}
\usepackage[dvipsnames]{xcolor}

\usepackage{multirow}
\usepackage{eqparbox}
\usepackage{makecell}

\usepackage{booktabs}
\usepackage{verbatim}

\newcommand{\green}[1] {\color[rgb]{0.13, 0.55, 0.13}(-\ensuremath #1\%)}

\definecolor{taborange}{RGB}{235, 127, 14}
\definecolor{tabblue}{RGB}{30, 144, 255}
\definecolor{tabgreen}{RGB}{30, 160, 30}

\usepackage{tikz}

\begin{document}

    \title{Weakly and Self-Supervised Class-Agnostic Motion Prediction for Autonomous Driving}
    \author{
	Ruibo Li,
	Hanyu Shi,
	Zhe Wang,
	Guosheng Lin
	\IEEEcompsocitemizethanks{
		\IEEEcompsocthanksitem Ruibo Li, Hanyu Shi, and Guosheng Lin are with the College of Computing and Data Science, Nanyang Technological University, Singapore (e-mail: ruibo001@e.ntu.edu.sg; hanyu001@e.ntu.edu.sg; gslin@ntu.edu.sg). 
		\IEEEcompsocthanksitem Zhe Wang is with SenseTime Research, Hong Kong, China (e-mail:wzlewis16@gmail.com).
	}   
	\thanks{
				Guosheng Lin is the corresponding author. 
	}
}

\IEEEpubid{\scriptsize 
	\textcopyright~2025 IEEE.
	This is the author’s accepted version. The final published paper is available at:
	\href{https://doi.org/10.1109/TPAMI.2025.3604036}{https://doi.org/10.1109/TPAMI.2025.3604036}}
\IEEEpubidadjcol

    \IEEEtitleabstractindextext{
        \begin{abstract}
Understanding motion in dynamic environments is critical for autonomous driving, thereby motivating research on class-agnostic motion prediction. In this work, we investigate weakly and self-supervised class-agnostic motion prediction from LiDAR point clouds. Outdoor scenes typically consist of mobile foregrounds and static backgrounds, allowing motion understanding to be associated with scene parsing. Based on this observation, we propose a novel weakly supervised paradigm that replaces motion annotations with fully or partially annotated (1\%, 0.1\%) foreground/background masks for supervision. To this end, we develop a weakly supervised approach utilizing foreground/background cues to guide the self-supervised learning of motion prediction models. Since foreground motion generally occurs in non-ground regions, non-ground/ground masks can serve as an alternative to foreground/background masks, further reducing annotation effort. Leveraging non-ground/ground cues, we propose two additional approaches: a weakly supervised method requiring fewer (0.01\%) foreground/background annotations, and a self-supervised method without annotations. Furthermore, we design a Robust Consistency-aware Chamfer Distance loss that incorporates multi-frame information and robust penalty functions to suppress outliers in self-supervised learning. Experiments show that our weakly and self-supervised models outperform existing self-supervised counterparts, and our weakly supervised models even rival some supervised ones. This demonstrates that our approaches effectively balance annotation effort and performance.

        \end{abstract}

    }

\maketitle

\begin{figure*}[ht]
	\centering 
	\includegraphics[width=0.90\textwidth]{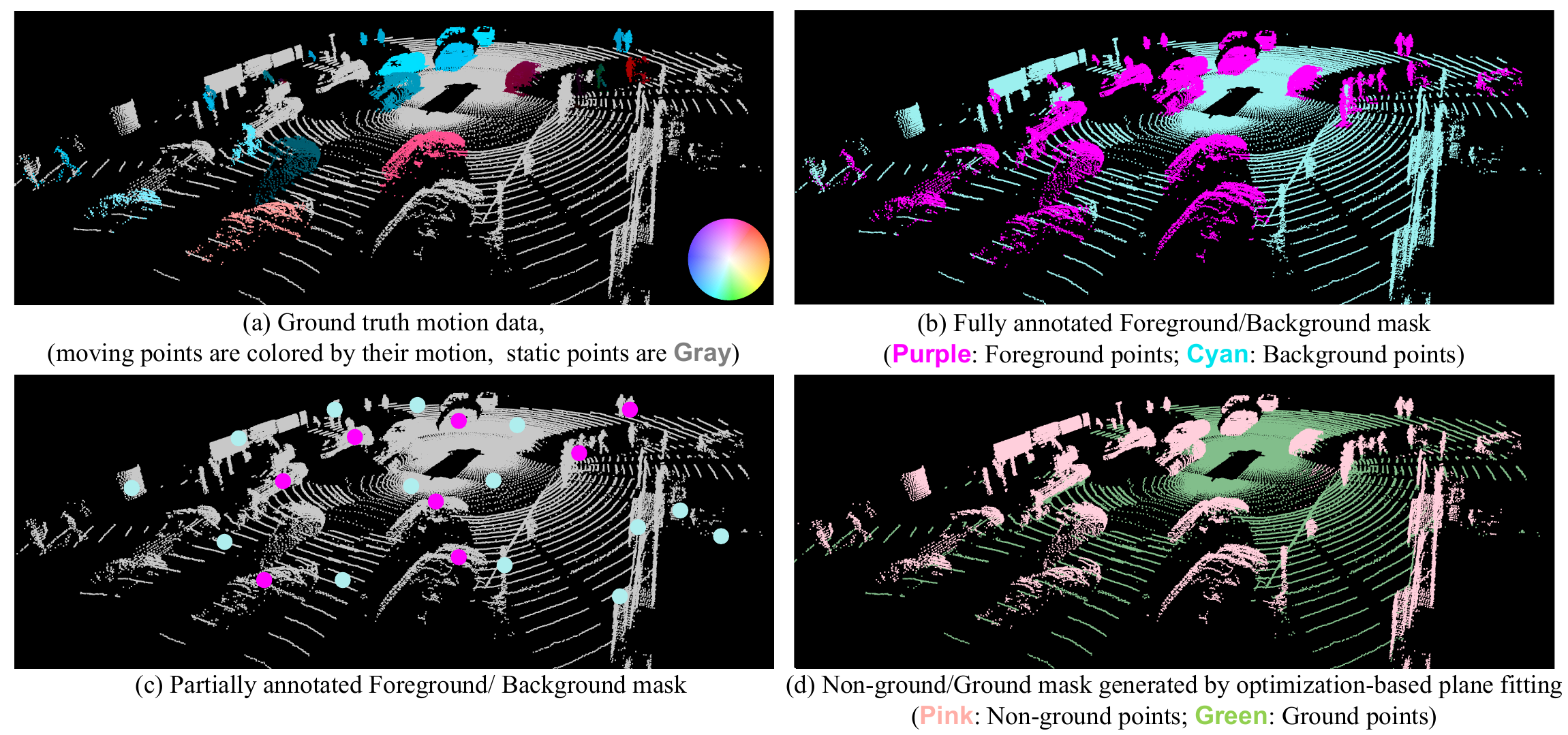}
	\vspace{-3mm}
	\caption{ Illustration of our weak and self-supervision concept.
		Outdoor scenes can be decomposed into mobile foreground and static background, which allows us to achieve motion learning with fully or partially annotated FG/BG masks as weak supervision, thereby replacing expensive motion annotations.
		The motion of foreground points typically occurs in non-ground areas, enabling us to use non-ground/ground masks as an alternative to FG/BG masks to further reduce or even eliminate annotation effort.
	}
	\label{fig_first}	
\end{figure*}

\section{Introduction}
\label{sec:intro}

Understanding  the dynamics of surrounding environments is vital for autonomous driving~\cite{menze2015object}.
In particular, motion prediction, which generates the future positions of objects based on past information, is crucial for ensuring safe planning and navigation.

Classical approaches~\cite{fang2020tpnet,zhao2021tnt,gao2020vectornet} employ object detection, tracking, and trajectory forecasting for motion prediction.
However, these detection-based approaches may fail when encountering unknown categories not present in the training data~\cite{wu2020motionnet}.
To address this issue, many methods~\cite{filatov2020any,wu2020motionnet,wang2022sti} suggest directly estimating class-agnostic motion from a bird’s eye view (BEV) map of point clouds, which achieves a good balance between accuracy and computational cost.
Nevertheless, sensors are unable to capture motion information in complex environments~\cite{menze2015object}, leading to a scarcity and high cost of motion data. 
Consequently, most existing real-world motion data are generated by semi-supervised learning methods with auxiliary information, such as KITTI~\cite{geiger2012we,menze2015object}, or derived from human-annotated object detection and tracking data, such as Waymo~\cite{jund2021scalable}.

To mitigate the dependence on motion annotations, many recent studies~\cite{wang2024semi,luo2021self,jia2023contrastmotion,wang2024self,fang2024self} investigate motion learning with little or no motion data.
Specifically, SSMP~\cite{wang2024semi} studies semi-supervised motion prediction using a small fraction of motion annotations for training. 
PillarMotion~\cite{luo2021self}, ContrastMotion~\cite{jia2023contrastmotion}, SSMotion~\cite{fang2024self}, and  SelfMotion~\cite{wang2024self} explore self-supervised motion prediction without relying on motion annotations for training.
Although these methods achieve promising results, there remains a significant performance gap compared to fully supervised methods.

Outdoor scenes can often be decomposed into a set of moving objects and backgrounds~\cite{menze2015object}, allowing us to associate motion understanding with scene parsing. 
As shown in Fig.~\ref{fig_first}~(a)~and~(b), with ego-motion compensation, motion only exists in foreground points.
Therefore,  if we distinguish the mobile foreground from the static background, we can focus on deriving valuable dynamic motion supervision from these potentially moving foreground objects, leading to more effective self-supervised motion learning.
Drawing from  this intuition, we study a novel weakly supervised paradigm, where costly motion annotations are replaced with fully or per-frame partially (\textbf{1\%}, \textbf{0.1\%}) annotated foreground/background (FG/BG) masks, achieving an effective balance between annotation effort and performance.
To this end, we propose \textbf{WeakMotion-FB}, a weakly supervised motion prediction approach guided by foreground/background cues.
Specifically,  we first train a FG/BG segmentation network using the partially annotated FG/BG masks, and then train a motion prediction network with guidance from the pretrained segmentation network.
Without motion annotations, the segmentation network generates moving foreground points for each training sample, from which dynamic motion supervision can be derived to enable self-supervised learning of the motion prediction network.
Additionally, an auxiliary FG/BG segmentation head is incorporated into the motion prediction network to regularize motion predictions by setting the motion in background regions to zero.

In outdoor scenes, the motion of foreground points typically occurs in non-ground areas, as illustrated in Fig.~\ref{fig_first}~(a),~(b),~and~(d).
This observation allows us to utilize non-ground masks as an alternative to foreground masks, enabling the extraction of dynamic motion supervision from the non-ground areas for self-supervised motion learning.
Building on this insight, and leveraging non-ground/ground cues, we propose two additional approaches: \textbf{WeakMotion-NG}, a weakly supervised method requiring much weaker FG/BG annotations, and \textbf{SelfMotion-NG}, a self-supervised method requiring no annotations.
Specifically, in  {SelfMotion-NG}, we segment non-ground points using optimization-based plane fitting, and train motion prediction networks on these non-ground points in a self-supervised manner.
In {WeakMotion-NG}, in addition to training the networks on the non-ground points,  we incorporate an auxiliary FG/BG segmentation head into  the networks for motion regularization.
Because WeakMotion-NG derives dynamic motion supervision from non-ground points rather than foreground points, it requires fewer FG/BG annotations compared to \mbox{WeakMotion-FB}, enabling training with much weaker supervision. 
With \mbox{WeakMotion-NG},  we achieve weakly supervised motion prediction by annotating 0.1\% of the points in a single frame of a 10-frame sequence. 
This new FG/BG annotation strategy further reduces the overall annotation effort to about \textbf{0.01\%}.

In self-supervised 3D motion learning~\cite{wu2020pointpwc,kittenplon2021flowstep3d,luo2021self}, the Chamfer distance (CD) loss function is preferred.
However, the CD is sensitive to outliers~\cite{tatarchenko2019single}.
Unfortunately, outliers are common in our setting.
This may arise from view changes and occlusions in point cloud sequences, as well as from potential errors in the FG points estimated by the FG/BG segmentation network and the non-ground points generated by plane fitting.
To alleviate the impact of outliers,  we propose a novel \textbf{R}obust \textbf{C}onsistency-aware \textbf{C}hamfer \textbf{D}istance (RCCD) loss.
Unlike the typical CD loss, our RCCD loss exploits supervision from multi-frame point clouds and leverages multi-frame consistency to measure the confidence of points.
By assigning uncertain points lower weights, our RCCD loss suppresses potential outliers.
Additionally, the RCCD loss adopts robust penalty functions, such as  the Geman-McClure penalty~\cite{geman1985,barron2019general}, to measure the distance between point clouds, making it more robust to outliers compared to the CD loss that employs the $L_2$-norm penalty.

Our main contributions can be summarized as follows:
	\begin{itemize}
	\item  
	Without using expensive motion data, we propose a  weakly supervised motion prediction paradigm with  fully or partially annotated foreground/background (FG/BG) masks as supervision, achieving a good compromise between annotation effort and performance.
	To the best of our knowledge, this is the first work on weakly supervised  class-agnostic motion prediction.
	
	\item  
	By associating motion understanding with scene parsing, we present a weakly supervised approach, {\textbf{WeakMotion-FB}}, which uses  FG/BG cues to guide the self-supervised learning of motion prediction models.
	This approach outperforms self-supervised methods by a large margin and performs on par with several fully supervised methods.
	
	\item 
	By using non-ground/ground masks as an alternative to FG/BG masks, we  develop another weakly supervised approach, \textbf{WeakMotion-NG}, which significantly reduces annotation effort while delivering satisfactory performance.
	
	\item
	We also introduce a self-supervised motion prediction approach, \textbf{SelfMotion-NG}, which requires no annotations and achieves state-of-the-art performance among self-supervised methods.
	
	\item 
	We design a novel Robust Consistency-aware Chamfer Distance loss, which leverages multi-frame information and robust penalty functions to suppress potential outliers for robust self-supervised motion learning.
	
\end{itemize}

The preliminary version of this work was presented in~\cite{li2023weakly}. 
We extend the conference version as follows: 
(1) By incorporating the previous Consistency-aware Chamfer Distance loss in~\cite{li2023weakly} with robust penalty functions, such as the Geman-McClure penalty, we design a novel Robust Consistency-aware Chamfer Distance loss to further improve the robustness to outliers.
Additionally, we add a smoothness regularization to the RCCD loss to promote local smoothness in motion predictions.
Experiments demonstrate that the two improvements significantly enhance the performance of models across all speed groups.
(2) We propose a new weakly supervised motion prediction approach (WeakMotion-NG) that further reduces the FG/BG annotation ratio by about 10$\times$ while providing satisfactory results.
(3)~By deriving motion supervision from non-ground points, we propose a new self-supervised  approach (SelfMotion-NG) with superior performance.
(4)~We present a more extensive overview of class-agnostic motion prediction and scene flow estimation. 
We have released the code and models of the preliminary version\footnote{https://github.com/L1bra1/WeakMotion}. The code and models for this extended version will be made available.

\section{Related Work}
\noindent\textbf{Class-agnostic motion prediction.} 
Classical approaches perform motion prediction by detecting potential traffic participants and estimating their future trajectories~\cite{fang2020tpnet,zhao2021tnt,gao2020vectornet,ettinger2021large,phan2020covernet,gu2021densetnt,zhang2020stinet,liang2020pnpnet,chang2019argoverse}.
However, since they rely on bounding box detection, their performance may degrade in open-set traffic scenarios, where the detectors may encounter object categories unseen during training and thus fail to recognize them.

In contrast to bounding box detection-based approaches, which are limited to predicting motion only for detected objects, many recent methods~\cite{filatov2020any,lee2020pillarflow,luo2021self,wu2020motionnet,wang2022sti,schreiber2021dynamic} represent the 3D environment using bird’s-eye view (BEV) maps derived from point clouds. This representation enables class-agnostic motion estimation for every region in the scene, rather than being constrained by predefined object categories.
PillarFlow~\cite{lee2020pillarflow} applies a 2D flow structure, PWC-Net~\cite{sun2018pwc}, to  establish correlations on BEV embeddings for motion estimation.
MotionNet ~\cite{wu2020motionnet} and  BE-STI~\cite{wang2022sti}  propose jointly predicting  semantic categories and future motion from BEV features.
In this work, we adopt the networks designed in MotionNet~\cite{wu2020motionnet} as our backbone modules.

To reduce the reliance on motion annotations, many recent works~\cite{wang2024semi,luo2021self,jia2023contrastmotion,wang2024self,fang2024self, li2024self} investigate motion prediction with little or no motion data.
Specifically, SSMP~\cite{wang2024semi} explores semi-supervised motion prediction using a small fraction of motion annotations for supervision.
PillarMotion~\cite{luo2021self} and SSMotion~\cite{fang2024self} propose a self-supervised learning framework that leverages point clouds and 2D optical flow information from camera images for self-training.
Additionally, without using any annotations or information from other modalities, RigidFlow++~\cite{li2024self} and SelfMotion~\cite{wang2024self} generate pseudo motion labels from point clouds for self-supervised learning. 
Despite these methods achieving promising results, there remains a significant performance gap compared to fully supervised methods.

Different from the above methods,  our work proposes a novel weakly supervised paradigm, where only fully or partially annotated foreground/background masks are used as weak supervision to balance annotation effort and performance.
Experimental results demonstrate that our weakly supervised models significantly outperform the self-supervised models.
In particular, our weakly supervised model, trained with 0.01\% FG/BG masks, surpasses the semi-supervised SSMP model~\cite{wang2024semi}, which is trained with  1\% motion annotations, by a large margin in both slow and fast speed groups.
Additionally, our work further proposes a self-supervised approach by using non-ground/ground masks as an alternative to FG/BG masks. This new approach also achieves superior performance compared to previous self-supervised methods.\\

\noindent\textbf{ Scene flow estimation.} 
Scene flow estimation~\cite{vedula1999three} focuses on computing the 3D motion field between two known point clouds.  Unlike motion prediction, which directly forecasts future motion from current observations, scene flow estimation computes the motion between the current and past frames. Under certain assumptions (e.g., linear dynamics), the computed scene flow can be extrapolated to predict future trajectories, as explored in MotionNet~\cite{wu2020motionnet}.
However, the high computational cost of most scene flow networks~\cite{liu2019flownet3d,cheng2022bi,li2021hcrf,li2022rigidflow,li2021self,wu2020pointpwc,gojcic2021weakly,dong2022exploiting,wang2022matters} hinders their applicability in real-time autonomous driving scenarios.

Weakly supervised scene flow estimation methods~\cite{gojcic2021weakly,dong2022exploiting} are related to our weakly supervised motion approaches.
However, our work differs from them in two key aspects:
(1) The purpose is different. The goal of our work is to forecast future motion based on past and current observations, while these methods focus on estimating current motion;
(2) Our supervision is much weaker than theirs. Our work can achieve weakly supervised motion learning with partially (1\%, 0.1\%, and 0.01\%) annotated FG/BG masks, whereas these methods rely on fully annotated masks.

Additionally, in weakly supervised and self-supervised scene flow estimation methods~\cite{mittal2020just, tishchenko2020self, kittenplon2021flowstep3d, baur2021slim, wu2020pointpwc, pontes2020scene, li2021self, gu2022rcp, he2022self, shen2023self, li2024self, li2022rigidflow, li2021neural, gojcic2021weakly,dong2022exploiting}, self-supervised motion loss functions and the generation of pseudo motion labels serve as the cornerstone.
Specifically, for self-supervised motion loss functions, \cite{mittal2020just, baur2021slim, tishchenko2020self} utilize the nearest neighbor loss, while \cite{wu2020pointpwc, kittenplon2021flowstep3d, pontes2020scene, gu2022rcp, shen2023self,li2021neural,gojcic2021weakly,dong2022exploiting} employ the Chamfer Distance loss.
Diverging from the nearest neighbor loss and the Chamfer Distance loss, our proposed Robust Consistency-aware Chamfer Distance loss incorporates supervision from multi-frame point clouds,  integrates robust penalty functions, and utilizes multi-frame consistency to suppress outliers, thereby achieving superior robustness.\\

\noindent\textbf{Self-supervised representation learning for point clouds.} 
To reduce reliance on motion annotations, the foundation of  our weakly and self-supervised motion prediction lies in self-supervised 3D motion learning for foreground or non-ground points. 
Similarly, motivated by the scarcity of manual annotations, self-supervised 3D representation learning has gained significant attention~\cite{xiao2023unsupervised}, aiming to learn meaningful point cloud representations from unlabeled data  and  transfer this knowledge to downstream applications, e.g., segmentation, detection, and classification. 
Most existing methods can be divided into two primary categories: contrastive learning-based methods~\cite{xie2020pointcontrast,sanghi2020info3d,zhang2021self,yin2022proposalcontrast,liu2022masked}, which learn representations by distinguishing one sample from others, and generation-based methods~\cite{yu2022point,pang2022masked,wang2021unsupervised,zhang2022point,huang2023ponder,chen2023pointgpt}, which learn representations by generating point clouds, such as through self-reconstruction or completion. 
In contrast, our self-supervised 3D motion learning is distinct from these methods in two fundamental aspects. (1) Objective: Our goal is to use self-supervision to directly address the motion prediction task itself, rather than using it as a pre-training strategy for other downstream tasks. (2) Source of supervision: We derive supervision signals by establishing temporal correspondences between points across consecutive frames and capturing the changes of their 3D coordinates over time, instead of relying on contrastive feature learning or point cloud generation.

\begin{figure}[htb]
	\centering 
	\includegraphics[width=0.49\textwidth]{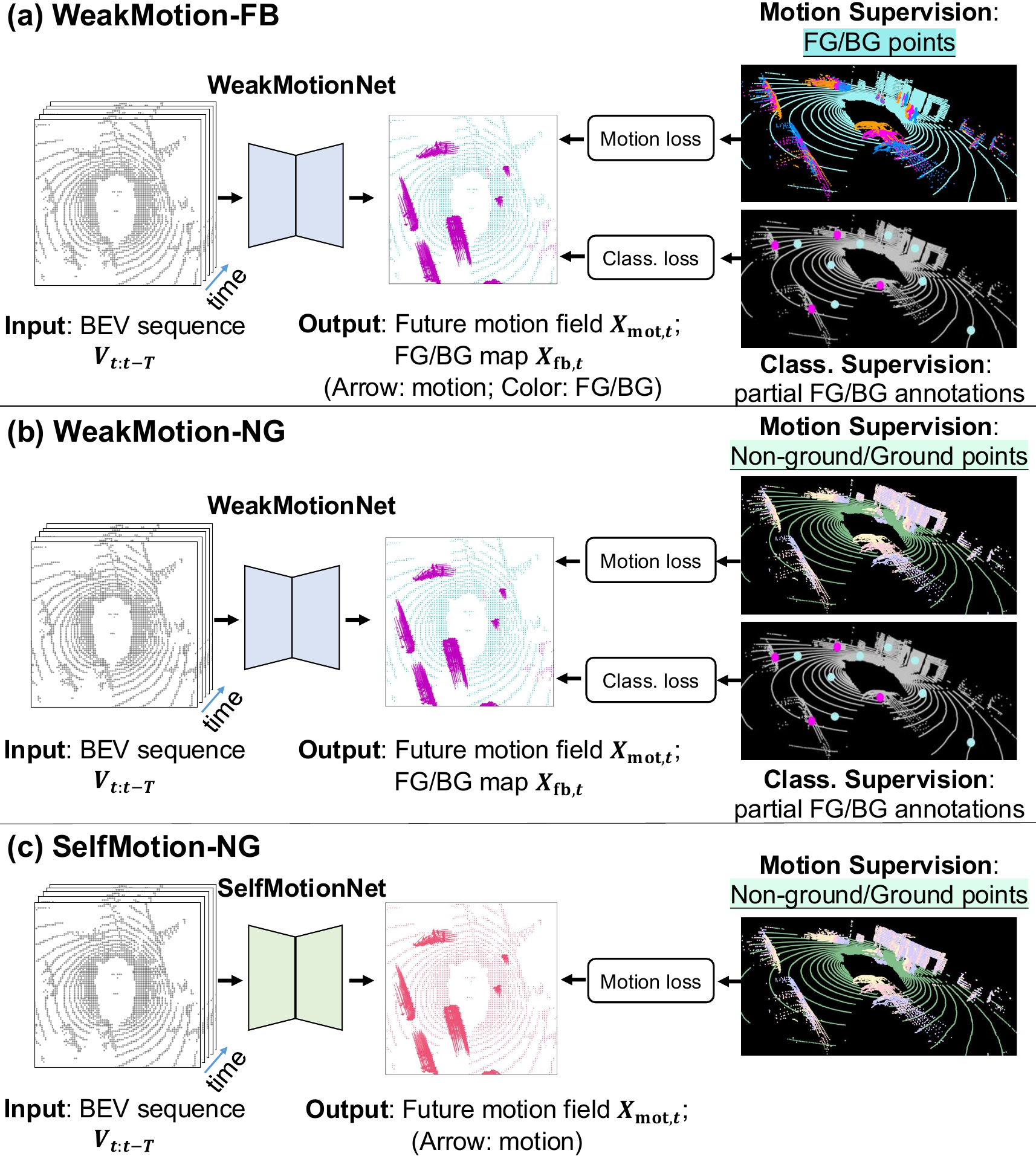}
	\vspace{-4mm}
	\caption{Summary of our proposed motion prediction approaches. (a) \textbf{WeakMotion-FB} trains a weakly supervised motion prediction model, WeakMotionNet, by deriving motion supervision from FG/BG points. (b) \textbf{WeakMotion-NG} trains WeakMotionNet by deriving motion supervision from Non-ground/Ground points. (C) \textbf{SelfMotion-NG} trains a self-supervised motion prediction model, SelfMotionNet, by deriving motion supervision from Non-ground/Ground points.}
	\label{sum}
\end{figure}

\section{Problem Formulation of Weakly and Self-supervised Motion Prediction}
For a sequence of consecutive LiDAR sweeps,  following previous works~\cite{wu2020motionnet,wang2022sti}, we first synchronize all point clouds to the current frame.
Specifically, we transform each past sweep into the current frame’s coordinate system using the ego-vehicle’s pose information, ensuring that all point clouds are spatially aligned to compensate for the ego-vehicle's motion.
Each synchronized point cloud at frame ${\tau}$ is denoted as ${\bm P_{\tau}} = \{{\bm p}_{\tau} (i) \in  \mathbb{R}^3 \}_{i=1}^{N_{\tau}}$, where $N_{\tau}$ is the number of points.
We then quantize ${\bm P_{\tau}}$ into a regular 3D voxel grid  ${\bm V_{\tau}} \in \{0,1\}^{H \times W \times C}$, where $0$ indicates an empty voxel and $1$ indicates an occupied one.
Here, $H$, $W$, and $C$ denote the number of voxels along the $X$, $Y$, and $Z$ axis, respectively.
By treating the $Z$~(vertical) axis as a feature dimension,  ${\bm V_{\tau}}$ can be regarded as a  2D bird’s eye view (BEV) map of size $H \times W $, 
where each cell  contains a binary vector of length~$C$ that encodes the height and occupancy information.

Taking the current and past BEV maps as input, the  class-agnostic motion prediction task aims to generate a future BEV motion field for the current frame~$t$.
The input is a sequence of BEV maps:
$\bm V_{t:t-T} = \{\bm V_{t},  \bm V_{t-1}, ..., \bm V_{t-T}\} \in \{0,1\}^{(T+1) \times H \times W \times C} $, where $T$ is the number of past frames used for prediction.
The output is a predicted future motion field ${\bm X_{{\rm  mot},t}}  \in  (\mathbb{R}^2)^{H \times W }$, where each vector $\bm x_{i,j}  \in  \mathbb{R}^2$ indicates the horizontal displacement of cell $(i,j)$ in the current BEV map $\bm V_{t}$ to its expected position at the next timestamp.
In our weakly supervised setting, we also estimate a foreground/background (FG/BG) category map $\bm X_{{\rm fb},t} \in (\mathbb{R}^2)^{H \times W}$ for the current BEV map $\bm V_{t}$ to regularize the motion prediction.

Furthermore, by assigning the motion of each cell to all points within this cell, we map ${\bm X_{{\rm  mot},t}}$ to the point level, resulting in per-point motion prediction ${\bm F_t} \in  \mathbb{R}^{N_t \times 3}$.
Specifically, since autonomous driving mainly focuses on the motion of dynamic agents along the ground plane, we set the vertical motion to zero.
This process is formulated as: 
\begin{equation}\label{eq_BEV_to_point}
{\bm F_t} = {\bm U_t}[{\bm X_{{\rm  mot},t}} ; { \vec{\bm 0}}],
\end{equation}
where $\bm U_t \in \{0,1\}^{N_t \times HW} $ is the assignment matrix derived from the spatial relationship  between the point cloud ${\bm P_t}$ and its corresponding BEV map ${\bm V_t}$.

Without relying on any motion ground truth, we investigate weakly and self-supervised motion prediction.
In our weakly supervised setting, we explore how to use fully or partially annotated FG/BG  masks for motion learning.
Specifically, for partially annotated FG/BG masks, we investigate two strategies:~\textbf{Partial annotation per frame}, where we randomly annotate 1\% or 0.1\% of the points in each frame;~\textbf{Partial annotation of a single frame per sequence}, where we sample  one frame from a 10-frame point cloud sequence and randomly annotate 0.1\% of the points in this sampled frame. 
Consequently,  the overall annotation ratio of the second strategy reaches about 0.01\%.  
In our self-supervised setting, we explore motion prediction without using any manual annotations.

 In  Sec.~\ref{chap_frame}, we introduce \textbf{WeakMotion-FB}, a weakly supervised motion prediction approach that utilizes FG/BG masks for motion supervision. 
 This approach supports both fully and per-frame partially (1\%, 0.1\%) annotated FG/BG masks.
Subsequently, in Sec.~\ref{chap_weaker}, we extend this approach by replacing FG/BG masks with non-ground/ground masks for motion supervision, enabling two alternative methods: (1) \textbf{WeakMotion-NG}, a weakly supervised approach designed for partial FG/BG annotations of a single frame per sequence, with an average annotation ratio of 0.01\% (Sec.~\ref{sec_weaker}), and (2) \textbf{SelfMotion-NG}, a self-supervised approach requiring no annotations (Sec.~\ref{sec_self}).
We summarize the proposed three approaches in Fig.~\ref{sum}.

\begin{figure*}[htb]
	\centering 
	\includegraphics[width=0.95\textwidth]{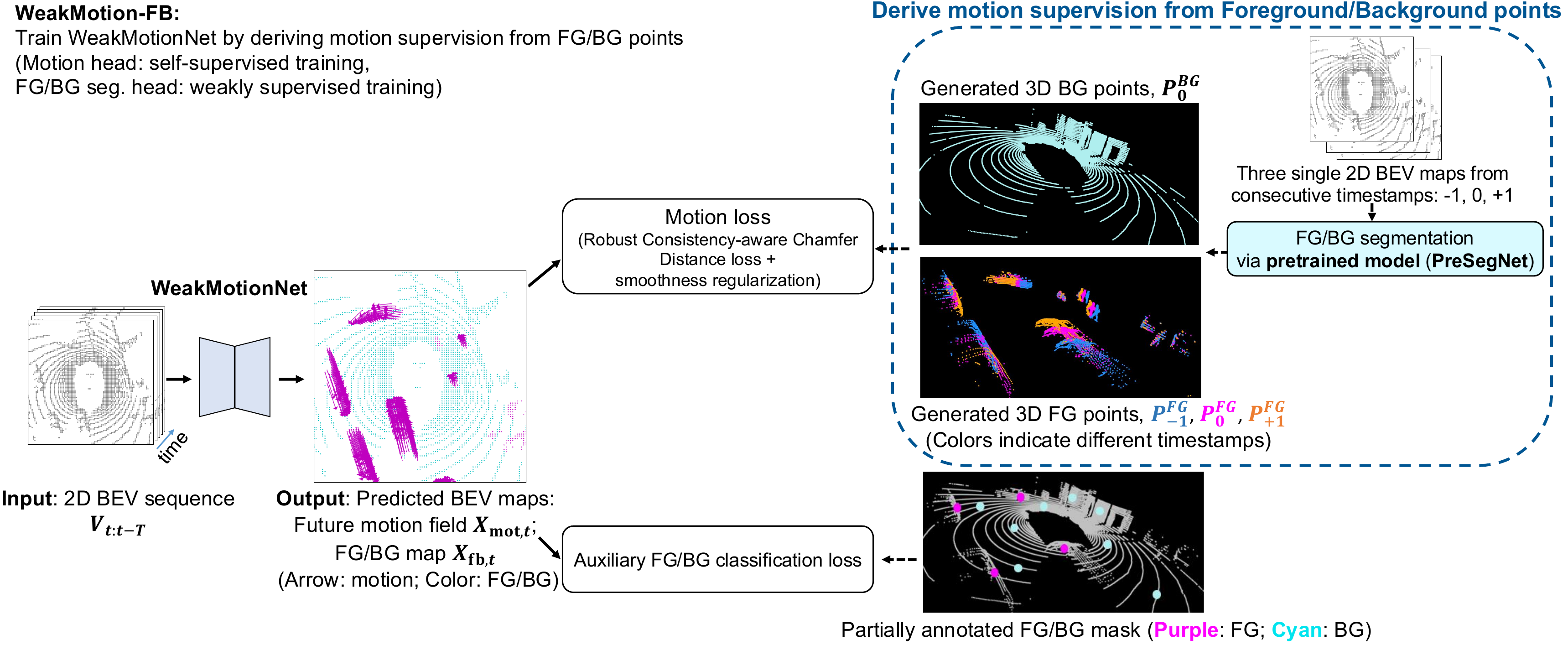}
	\caption{ 
		Overview of \textbf{WeakMotion-FB}, a weakly supervised motion prediction approach guided by \textbf{foreground/background cues}.
		In \mbox{{WeakMotion-FB}}, we train a motion prediction network, \textbf{WeakMotionNet}, to predict future motion field~${\bm X_{{\rm  mot},t}}$ and FG/BG category map~${\bm X_{{\rm  fb},t}}$ from a sequence of synchronized BEV maps $\bm V_{t:t-T}$.
		To derive motion supervision, we use a pretrained segmentation model (\textbf{PreSegNet}, detailed in Fig.~\ref{fig_preseg}) to generate FG/BG points, and apply a Robust Consistency-aware Chamfer Distance loss to these points, enabling self-supervised learning of the motion prediction head. Meanwhile, the auxiliary FG/BG segmentation head is trained with partially annotated FG/BG masks.
}
	\label{fig_pipeline}
\end{figure*}

\begin{figure*}[htb]
	\centering 
	\includegraphics[width=0.95\textwidth]{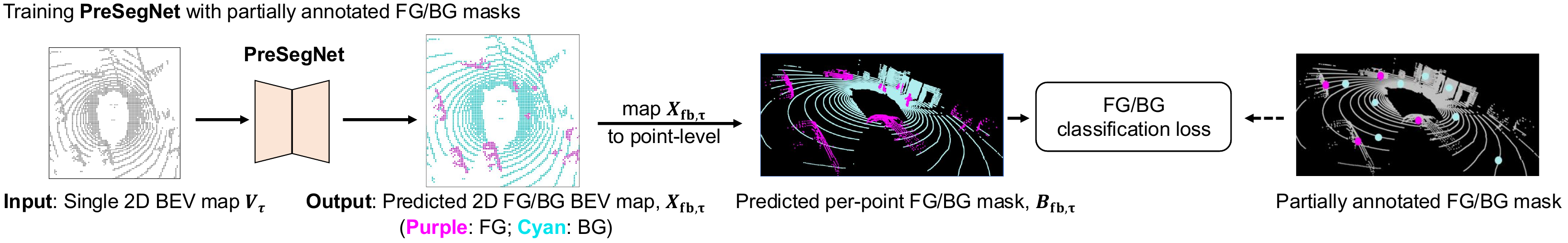}
	\caption{Overview of \textbf{PreSegNet}. 
		Given partially annotated FG/BG masks, we train a foreground/background (FG/BG) segmentation network (\textbf{PreSegNet}), which provides complete per-point FG/BG masks to support the self-supervised motion learning in \textbf{WeakMotion-FB}.}
	\label{fig_preseg}
\end{figure*}

\section{Motion prediction with foreground and background cues}\label{chap_frame}

The weakly supervised motion prediction approach, \textbf{WeakMotion-FB}, illustrated in Fig.~\ref{fig_pipeline}, is designed for fully and per-frame partially annotated FG/BG masks.

To leverage FG/BG cues for guiding motion learning, we first train a  FG/BG segmentation network, \textbf{PreSegNet}, using per-frame partially annotated FG/BG masks as weak supervision, as depicted in Fig.~\ref{fig_preseg}. 
Subsequently, in WeakMotion-FB, we train a motion prediction network, \textbf{WeakMotionNet}, with two output heads: a motion prediction head and an auxiliary FG/BG segmentation head.
In the training of motion prediction head, for each training sample $\bm V_{t:t-T}$, we first select three consecutive point clouds from the past (-1), current (0) and future (+1) timestamps.
And then, we use the trained PreSegNet to generate FG/BG points for the three frames. 
Based on these generated FG/BG points, we employ a novel Robust Consistency-aware Chamfer Distance loss function and a smoothness regularization for motion learning.
In the training of auxiliary FG/BG segmentation head, we also adopt per-frame partially annotated FG/BG masks as weak supervision.
The training of both heads in WeakMotionNet is performed simultaneously.
Specifically, when fully annotated FG/BG masks are available, PreSegNet can be omitted, and the complete FG/BG labels can be directly used to generate 3D FG/BG points for motion learning.

In this section, we first revisit the Chamfer loss in 3D motion tasks (Sec.~\ref{Preliminaries}) and then discuss the details about our proposed Robust Consistency-aware Chamfer Distance loss (Sec.~\ref{CCL}).
Finally, we will introduce the architecture of the two networks and present their training strategies  (Sec.~\ref{NI}).

\subsection{Preliminaries:  Chamfer Loss in 3D Motion}\label{Preliminaries}
Chamfer Distance (CD) is widely used in various point cloud tasks, such as completion~\cite{wu2021density}, generation~\cite{yang2019pointflow}, reconstruction~\cite{choe2021deep}, and 3D motion perception~\cite{wu2020pointpwc,kittenplon2021flowstep3d,luo2021self,pontes2020scene,dong2022exploiting,gu2022rcp,wang2022neural,li2021neural,ouyang2021occlusion,he2022learning}.
Given two consecutive point sets $\bm {S_1}$ and $\bm {S_2}$, and the predicted per-point motion $\bm {F}$ from models, the Chamfer Distance loss for self-supervised 3D motion learning can be defined as:
\begin{equation}\label{CD_loss}\small
\begin{aligned}
{ \bm {\widehat S_1}} =  & \bm {S_1} + \bm {F},\\
{\cal L}_{CD}(\bm {\widehat S_1}, \bm {S_2}) \!=  & \frac{1}{|\bm {\widehat S_1}|}\!\sum_{ \bm {x}\in \bm {\widehat S_1}}\!\min_{\bm {y}\in \bm {S_2}}\!|| \bm {x} \!-\! \bm {y} ||_2^2 
\!+\! \frac{1}{|\bm {S_2}|}\!\sum_{\bm {y} \in \bm {S_2}}\!\min_{\bm {x} \in \bm {\widehat S_1}}\!|| \bm {y} \!-\! \bm {x}||_2^2, 
\end{aligned}
\end{equation}
where  ${ \bm {\widehat S_1}} $ is the warped first point set using the predicted motion.
By minimizing the CD between ${ \bm {\widehat S_1}} $ and ${ \bm {S_2}}$, the models learn to predict the motion that moves the first point set  to the second set.

\subsection{Robust Consistency-aware Chamfer Distance Loss}\label{CCL}
In motion prediction, point clouds are synchronized, and motion only exists in foreground points.
Therefore, in WeakMotion-FB, we use the pretrained  PreSegNet  to generate possible foreground (FG) and background (BG) points of training samples, and train the motion prediction head of WeakMotionNet on the potentially moving foreground points for more effective self-supervised motion learning.

Chamfer Distance (CD) loss is a viable option for self-supervised motion learning; however, it is sensitive to outliers~\cite{tatarchenko2019single}. 
In this task, outliers are quite common due to view-changes and occlusions in the point clouds, as well as potential errors in the generated FG points.
To alleviate the impact of outliers,  we propose a novel  \textbf{R}obust \textbf{C}onsistency-aware \textbf{C}hamfer \textbf{D}istance (RCCD) loss function.
Compared to the original CD loss (Eq.~(\ref{CD_loss})),  Our RCCD loss is improved in three aspects.
(1) Our RCCD loss minimizes not only the distance between the forward warped current data and the future data, but also the distance between the backward warped current data and the past data, thus deriving  supervision from multi-frame information.
(2) Our RCCD loss employs multi-frame consistency to measure the confidence of points and assigns uncertain points less weight to suppress potential outliers. 
(3) Our RCCD loss can adopt various robust penalty functions (e.g., the $L_1$-norm, the Welsch-Leclerc penalty~\cite{dennis1978techniques,leclerc1989constructing}, and the Geman-McClure penalty~\cite{geman1985,barron2019general}) to calculate the distance between two point clouds, making RCCD loss more robust to outliers than the original CD loss that employs the $L_2$-norm penalty.

For each training sample, the generated foreground points from the past (-1), current (0), and future (+1) timestamps are denoted as ${\bm P_{-1}^{\rm FG}}, {\bm P_{0}^{\rm FG}}, {\bm P_{+1}^{\rm FG}}$, respectively.
The predicted motion of the generated foreground points in the current timestamp is denoted as ${\bm F_{0}^{\rm FG}}$.
The per-point motion ${\bm F^{\rm FG}}$ is obtained by mapping the predicted BEV motion field ${\bm X_{{\rm  mot}}}$  to the point level.
For simplicity, we omit the $\bm { FG}$ superscript in this subsection.\\

\begin{figure}[tb]
	\centering 
	\includegraphics[width=0.45\textwidth]{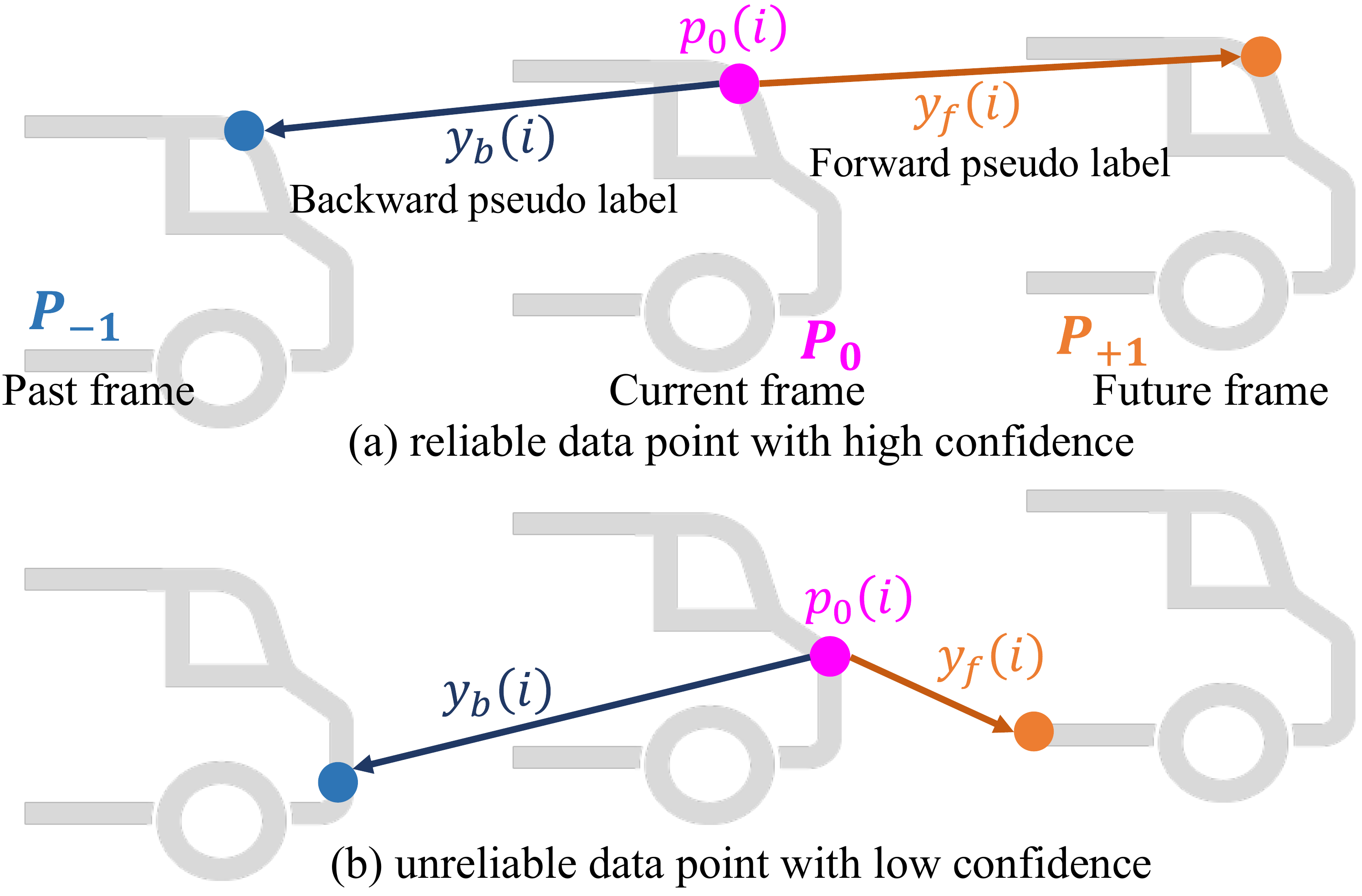} 
	\vspace{-2mm}
	\caption{Illustration of confidence estimation. 
		Assuming that an object’s motion remains consistent within a short temporal window, a reliable data point should  have  forward and backward pseudo motion labels that are equal in magnitude but opposite in direction.
		Therefore, we measure the confidence of each data point~$ {\bm { p}}_{0}(i)$  based on the consistency between its forward pseudo label ${\bm y}_{f}(i)$ and backward pseudo label ${\bm y}_{b}(i)$. 
		(a) A reliable data point with high confidence, where the forward and backward pseudo motion labels are nearly symmetric.
		(b) An unreliable data point with low confidence, where the forward and backward pseudo motion labels are inconsistent.}
	\label{fig_confidence}	
\end{figure}

\noindent\textbf{Warping the predicted foreground points.}\quad 
Assuming the motion of objects is consistent within a short temporal window, 
we obtain the forward warped  FG points in current frame ${\bm {\widehat P}}_{0,f}$ by warping the predicted current FG points ${\bm P_{0}}$ with their predicted motion ${\bm F_0}$, and obtain the backward warped  FG points  ${\bm {\widehat P}}_{0,b}$ by warping ${\bm P_{0}}$ with the inverse of their predicted motion $\text{-} {\bm F_0}$:
\begin{equation}
\begin{aligned}
{\bm {\widehat P}}_{0,f} = {\bm P_0} + {\bm F_0},  \quad {\bm {\widehat P}}_{0,b} = {\bm P_0} - {\bm F_0}.
\end{aligned}
\end{equation}

\noindent\textbf{Estimating the confidence of points.}\quad 
The CD loss minimizes the distance between the warped current data and the future data.
For each point, the CD loss finds its closest point in the other point cloud as correspondence and uses the coordinate difference as a pseudo label to approximate motion ground truth of this point. 

A reliable data point should have a consistent  pseudo motion label within a short time window.
Based on this intuition, given point clouds from three consecutive timestamps, our RCCD loss generates forward and backward pseudo labels and uses  the consistency to measure the confidence of this point. 
An example is shown in Fig.~\ref{fig_confidence}.
By reweighting our loss function with the confidence, data points with consistent pseudo labels dominate the training, while outliers are suppressed.	
Table~\ref{tab_CCD} shows the effectiveness of the confidence reweighting and Fig.~\ref{fig_chamfer} provides a visualization example.
The confidence generation for each point in ${\bm P_{0}}$ can be formulated as follows:
	\begin{align}
	{\bm y}_{f}(i) & =  \arg \min_{\bm {s}\in \bm {P}_{\text{+}1}} \|  \bm {s} - {\bm {\widehat p}}_{0,f}(i)  \|_2 - {\bm { p}}_{0}(i), \label{eq_forward_pseudo_label}\\
	{\bm y}_{b}(i) & =   \arg \min_{\bm {s}\in \bm {P_{\text{--}1}}} \|  \bm {s} - {\bm {\widehat p}}_{0,b}(i)   \|_2 - {\bm { p}}_{0}(i), \label{eq_backward_pseudo_label}\\
	w_0(i) & =\exp(\frac{-  \| {\bm y}_{f}(i) +  {\bm y}_{b}(i) \|_2^2}{2 \theta^2}).\label{eq_Gaussian_pseudo_label}
	\end{align}
In Eq.~(\ref{eq_forward_pseudo_label}), for a point $ {\bm { p}}_{0}(i)$ in ${\bm P_{0}}$, we find the closest point in   ${\bm P_{\text{+}1}}$ to ${\bm {\widehat p}}_{0,f}(i)$ as the correspondence of $ {\bm { p}}_{0}(i)$, and take the coordinate difference as  the forward  pseudo label ${\bm y}_{f}(i)$. 
Based on the same strategy, we also obtain its backward  pseudo label ${\bm y}_{b}(i)$ in Eq.~(\ref{eq_backward_pseudo_label}). 
After that, by taking the consistency between ${\bm y}_{f}(i)$ and ${\bm y}_{b}(i)$  as a metric, we use a Gaussian kernel to generate its confidence score $w_0(i)$  in Eq.~(\ref{eq_Gaussian_pseudo_label}).
In our experiments, we set $\theta^2$ to $0.5$.

According to the confidence map ${\bm { w}}_{0}$ for ${\bm P_{0}}$, the confidence map for ${\bm P_{\text{+}1}}$ and ${\bm P_{\text{--}1}}$ can be obtained by  nearest search.
Here is an example of generating confidence score for a point $ {\bm { p}}_{\text{+}1}(j)$ in ${\bm P_{\text{+}1}}$:
\begin{align}
	I_{\text{+}1}(j) & =  \arg \min_{i \in \{1, ..., N_0 \} } \| {\bm { p}}_{\text{+}1}(j)  -{\bm {\widehat p}}_{0,f}(i)  \|_2, \label{eq_index}\\
	w_{\text{+}1}(j) & = w_0(I_{\text{+}1}(j)). \label{eq_nn_confidence}
\end{align}
In Eq.~(\ref{eq_index}), for a  point ${\bm { p}}_{\text{+}1}(j) $ in ${\bm P_{\text{+}1}}$, we find its closest point in ${\bm {\widehat P}}_{0,f}$ and get the index of this closest point, $I_{\text{+}1}(j)$. 
Then, in Eq.~(\ref{eq_nn_confidence}), we take the confidence score of its closest point as the confidence score $w_{\text{+}1}(j)$ for ${\bm { p}}_{\text{+}1}(j) $.\\

\noindent\textbf{RCCD loss function with various robust penalties. }\quad 
The Robust Consistency-aware Chamfer Distance (RCCD) loss function can be written as:
\begin{equation}\label{CCD}
\begin{aligned}
{\cal L}_{RCCD}({\bm P_{\text{--}1}}, {\bm P_{0}}, {\bm P_{\text{+}1}}, {\bm F_0}) =
{\cal L}_{SR}({\bm {\widehat P}}_{0,b}, {\bm { P}}_{\text{--}1},  {\bm { w}}_{0}, {\bm { w}}_{\text{--}1}) \\
+ {\cal L}_{SR}({\bm {\widehat P}}_{0,f}, {\bm { P}}_{\text{+}1},  {\bm { w}}_{0}, {\bm { w}}_{\text{+}1}), 
\end{aligned}
\end{equation}
where the first term minimizes the distance between the backward warped current points and the past points, and the second term minimizes the distance between the forward warped current points and the future points.
Taking the second term as an example, the $L_{SR}$ can be formulated as: 
\begin{equation}\small
\begin{aligned}
{\cal L}_{SR}({\bm {\widehat P}}_{0,f}, \!{\bm { P}}_{\text{+}1}, \!{\bm { w}}_{0}, \!{\bm { w}}_{\text{+}1})  
\! = \! \frac{1}{  \| {\bm w}_0 \|_1 }\!\sum_{ i =1}^{N_0}  \!w_0(i) \!\min_{\bm {s}\in {\bm { P}}_{\text{+}1}} \!   \rho \left({\bm {\widehat p}}_{0,f}(i) - \bm {s} \right)  \\
+\frac{1}{ \| {\bm w}_{\text{+}1} \|_1 }\sum_{ j =1}^{N_{\text{+}1}}  w_{\text{+}1}(j) \!\min_{\bm {s}\in {\bm {\widehat P}}_{0,f} } \rho \left( { \bm { p}}_{\text{+}1}(j) - {\bm s}\right), 
\end{aligned}
\end{equation}
where $ \rho \left(  \cdot \right)$ is the robust penalty function.

In this paper, we combine our RCCD loss with various robust penalties, including $L_1$-norm penalty $ \rho (\bm {\epsilon}) = \| \bm {\epsilon} \|_1$,  Welsch-Leclerc penalty $ \rho (\bm {\epsilon}) = 1 - \exp{(- {\bm {\epsilon}}^2 /2 )}$, and Geman-McClure penalty $\rho (\bm {\epsilon}) = \frac{{\bm {\epsilon}^2}}{{\bm {\epsilon}}^2 + 1}$.
Therefore, ${\cal L}_{SR}$ can be viewed as a weighted Chamfer loss, with the confidence map serving as weights to suppress potential outliers and the robust penalty function as a metric to measure distances. 
Specifically, we use  Geman-McClure penalty as the default robust penalty function of our RCCD loss.\\

\noindent\textbf{Spatial smoothness regularization.}\quad 
Inspired by self-supervised scene flow methods~\cite{wu2020pointpwc,kittenplon2021flowstep3d}, we incorporate a smoothness regularization into the RCCD loss to encourage local smoothness in motion predictions:
\begin{equation}\label{eq.smooth}
{\cal L}_{\rm smooth} ({\bm F_0}) = \frac{1}{ N_0}\sum_{ i =1}^{N_0} \frac{1}{ | {\cal C}_i |}  \sum_{ j \in {\cal C}_i }  \|  {\bm f_0}(i) -{\bm f_0}(j)\|_1 ,
\end{equation}
where $N_0$ is the point number of the predicted current FG points,  
${\bm F_0}$ is the predicted motion of the FG points, 
and $ {\cal C}_i $ is a set of neighboring FG points around each FG point, ${\bm p_{0}}(i)$.\\

\subsection{Network Implementation}\label{NI}

\subsubsection{Pre-segmentation Network (PreSegNet)}\label{PreSegNet}

PreSegNet is a foreground/background (FG/BG) segmentation model comprising a backbone network and a FG/BG segmentation head. The backbone network is based on the structure used in MotionNet~\cite{wu2020motionnet}, modified by removing the temporal convolutions from each block to adapt it for single-frame segmentation. The FG/BG segmentation head is a two-layer 2D convolutional network.\\ 

\noindent\textbf{ Training.}\quad 
As depicted in Fig.~\ref{fig_preseg}, for each frame $\tau$, the point cloud ${\bm P_{\tau}} $ is quantized into a single BEV map ${\bm V_{\tau}} $.
PreSegNet takes ${\bm V_{\tau}} $ as input and predicts its FG/BG category map ${\bm X_{{\rm  fb},\tau}}$.
In our weakly supervised setting, the FG/BG labels are only available in a small fraction of points in  ${\bm P_{\tau}} $.
To train PreSegNet with incomplete point-wise supervision, we first map ${\bm X_{{\rm  fb},\tau}}$ to the point level to obtain per-point category predictions ${\bm B_{{\rm  fb},\tau}} $. 
This process is formulated as: ${\bm B_{{\rm  fb},\tau}} = {\bm U_\tau}{\bm X_{{\rm  fb},\tau}}$, where ${\bm U_\tau}$ is a $0\text{--}1$ assignment matrix derived from the spatial relationship  between ${\bm P_{\tau}}$ and ${\bm V_{\tau}}$.
Then, the FG/BG classification loss  can be written as:
\begin{equation}\label{cls}
{\cal L}_{\rm cls} = \frac{1}{ | \cal{D}_{\tau} |} \sum_{i \in \cal{D}_{\tau}} \alpha_{\tau}(i)  \cdot {\rm  CE}({\bm b_{{\rm  fb},\tau}} (i), {{\bm b^{\rm gt} _{{\rm  fb},\tau}}} (i)),
\end{equation}
where $\cal{D}_{\tau}$ is the set of labeled points in ${\bm P_{\tau}}$, ${\rm  CE}(\cdot)$ is a cross-entropy loss, ${\bm b_{{\rm  fb},\tau}} (i)$ is the predicted FG/BG category of point $i$, and ${\bm b^{\rm gt}_{{\rm  fb},\tau}} (i)$  is its label.
Specifically, $\alpha_{\tau}(i)$ is the weight assigned to different categories.

\subsubsection{Motion Prediction Network (WeakMotionNet)}
WeakMotionNet comprises a motion prediction network containing a backbone network, a motion prediction head, and an auxiliary FG/BG segmentation head.
We implement the backbone network using the same structure as the one in MotionNet~\cite{wu2020motionnet}  and implement the two output heads as two-layer 2D convolutional networks.\\

\noindent\textbf{ Training.}\quad 
As illustrated in Fig.~\ref{fig_pipeline}, in each frame $t$, the WeakMotionNet takes a sequence of synchronized BEV maps $\bm V_{t:t-T}$ as input and predicts the future motion map ${\bm X_{{\rm  mot},t}}$ and  FG/BG category map ${\bm X_{{\rm  fb},t}}$ of frame $t$.
Using the assignment matrix  ${\bm U_t}$, we get the point-wise motion  ${\bm F_t}$ and category ${\bm B_{{\rm  fb},t}}$, as presented in Eq.~(\ref{eq_BEV_to_point}).

In the training of motion prediction head, we select three consecutive point clouds from the past $(t\text{-}1)$, current $(t)$, and future $(t\text{+}1)$ timestamps.
We then use the pretrained PreSegNet  to generate their corresponding FG/BG points. 
Specially, in our experiments, we set the time span between two timestamps to 0.5s.
Since the point clouds are synchronized, we consider the generated BG points as static and apply the RCCD loss and the  regularization exclusively to the generated FG points, ${\bm P_{t\text{-}1}^{\rm FG}}, {\bm P_{t}^{\rm FG}}, {\bm P_{t\text{+}1}^{\rm FG}}$.
Therefore, the motion loss contains three parts:
\begin{equation}\small
\begin{aligned}
{\cal L}_{\rm mot}  = &{\cal L}_{RCCD}({\bm P^{\rm FG}_{t\text{-}1}}, {\bm P^{\rm FG}_{t}}, {\bm P^{\rm FG}_{t\text{+}1}}, {\bm F_{t}^{\rm FG}}) + {\cal L}_{\rm smooth} ({\bm F_{t}^{\rm FG}})  \\
 & +  \varphi_{bg} {\cal L}_{\rm mot,S} ({\bm F_t^{\rm BG}}) 
\end{aligned}
\end{equation}
where the first term is the RCCD loss (Eq.~(\ref{CCD})) for the predicted motion of the generated FG points; ${\bm F_t^{\rm FG}}$,  the second term is the smoothness regularization (Eq.~(\ref{eq.smooth})) for ${\bm F_t^{\rm FG}}$; and the third term is for the predicted motion of the generated BG points, ${\bm F_t^{\rm BG}}$.  
$\varphi_{bg}$~is the weight coefficient for the motion loss of the generated BG points.
Regarding the generated BG points as static, we train their predicted motion, ${\bm F_t^{\rm BG}}$, to be zero:
\begin{equation}\label{eq.mot_S}
\begin{aligned}
{\cal L}_{\rm mot,S} ({\bm F_t^{\rm BG}}) = \frac{1}{ N^{\rm BG}_t}\sum_{ i =1}^{N^{\rm BG}_t} \|  {\bm f^{\rm BG}}(i) -{ \vec{\bm 0}}\|_1,
\end{aligned}
\end{equation}
where $N^{\rm BG}_t$ is the number of the generated BG points.

In the training of auxiliary FG/BG segmentation head, we use the same classification loss (Eq.~\ref{cls}) and follow the same strategy used in Sec.~\ref{PreSegNet}.
The total loss in WeakMotion-FB is the combination of the two loss functions: 
\begin{equation}
{\cal L}_{\rm wfb} = \beta_1 {\cal L}_{\rm cls} + \beta_2 {\cal L}_{\rm mot},
\end{equation}
where $\beta_1$ and $\beta_2$ are the weight coefficients for different loss terms.
Specifically, we set $\beta_1$ to $1$ and $\beta_2$ to $5$.
In  inference, we regularize the final motion predictions by setting the motion of predicted background areas to zero.

\begin{figure*}[htb]
	\centering 
	\includegraphics[width=0.95\textwidth]{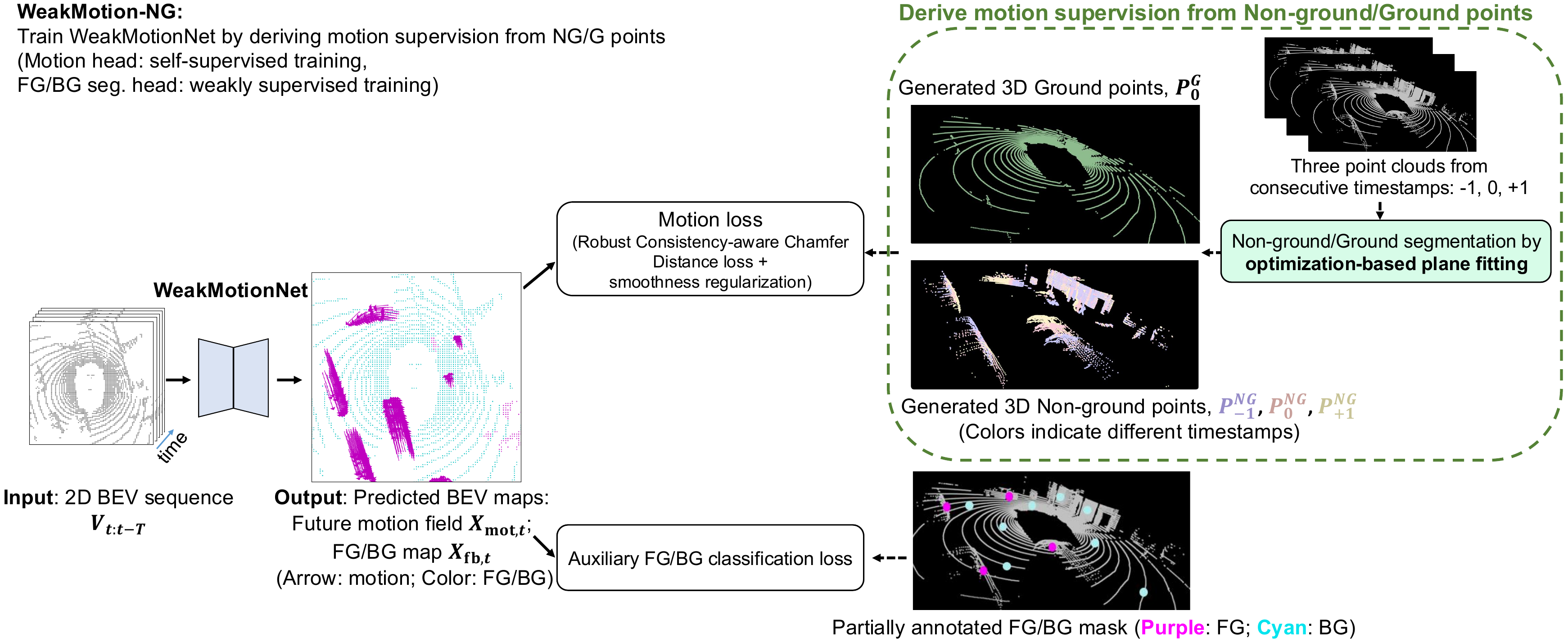}
	\vspace{-2mm}
	\caption{
		Overview of \textbf{WeakMotion-NG}, a weakly supervised motion prediction approach guided by \textbf{non-ground/ground cues}.
		In {WeakMotion-NG}, we train a motion prediction network, \textbf{WeakMotionNet}, to predict future motion field~${\bm X_{{\rm  mot},t}}$ and FG/BG category map~${\bm X_{{\rm  fb},t}}$ from a sequence of synchronized BEV maps $\bm V_{t:t-T}$.
		To derive motion supervision, we use an \textbf{optimization-based plane fitting} to generate non-ground/ground points, and apply a Robust Consistency-aware Chamfer Distance loss to these points, enabling self-supervised learning of the motion prediction head.		
		 Meanwhile, the auxiliary FG/BG segmentation head is trained with partially annotated FG/BG masks.}
	\label{fig_WeakPP_pipeline}
\end{figure*}

\section{Motion prediction with non-ground and ground cues}\label{chap_weaker}
The motion of foregrounds typically occurs in non-ground areas, allowing us to utilize non-ground/ground masks as an alternative to FG/BG masks to further reduce or even eliminate annotation effort.
In Sec.~\ref{sec_weaker}, we introduce  \textbf{WeakMotion-NG}, a weakly supervised approach requiring much weaker FG/BG annotations, i.e., partial FG/BG annotations for a single frame in each sequence.
Subsequently,  in Sec.~\ref{sec_self}, we present \textbf{SelfMotion-NG}, a self-supervised motion prediction approach without using  annotations.

\subsection{Motion prediction with much weaker FG/BG supervision}\label{sec_weaker}

When we partially annotate only one frame per sequence, a straightforward approach to enable weakly supervised motion prediction is to apply WeakMotion-FB proposed in Sec.~\ref{chap_frame}.
However, training with such weak FG/BG annotations may degrade the performance of the FG/BG segmentation model, PreSegNet, which may consequently fail to produce accurate FG/BG points, ultimately impairing motion learning in WeakMotion-FB.

To address this issue, we develop \textbf{WeakMotion-NG}, a weakly supervised motion prediction approach guided by non-ground/ground cues.
As shown in Fig.~\ref{fig_WeakPP_pipeline}, we generate non-ground and ground points by an optimization-based ground segmentation and employ non-ground/ground masks as an alternative to FG/BG masks. 
Leveraging the ground segmentation results, we apply the Robust Consistency-aware Chamfer Distance loss to the non-ground points, thereby enabling the direct extraction of dynamic motion supervision from these non-ground areas without relying on the pretrained FG/BG segmentation model.
In parallel, FG/BG annotations are used solely to train the auxiliary FG/BG segmentation head in WeakMotionNet. 
Therefore, compared to  \mbox{WeakMotion-FB}, WeakMotion-NG reduces dependency on FG/BG annotations, allowing WeakMotionNet to be trained with substantially weaker FG/BG supervision.

In the following, we introduce ground segmentation using RANSAC-based plane fitting, followed by the implementation details of WeakMotion-NG.

\subsubsection{Ground Segmentation by RANSAC-based Plane Fitting}\label{chap_ground_seg}
For each point cloud  ${\bm P}$, we generate the non-ground points ${\bm P^{NG}}$ and ground points ${\bm P^{G}}$ by plane fitting.
Specifically, inspired from scene flow methods~\cite{najibi2022motion,chodosh2023re}, we apply the RANSAC algorithm~\cite{fischler1981random} to plane fitting.

Firstly, we extract points with lower heights from ${\bm P}$ as candidate points.
In each iteration of RANSAC, we randomly sample three points from the candidate points and then estimate a plane to fit the three points.
Only planes that are nearly horizontal will be considered as valid.
We then compute the Euclidean distance from the candidate points to the valid plane, treating any point with a distance lower than a threshold, $d_{thresh}$, as an inlier.
After all RANSAC iterations, the plane with the maximum inliers is chosen as the optimal fit. 
Subsequently, we compute the Euclidean distance from all points in ${\bm P}$ to this optimal plane.
Points with distances less than $d_{thresh}$ are categorized as ground, while those with greater distances are classified as non-ground.
Specifically, we set $d_{thresh}$ to $0.4$m.

\begin{figure*}[tb] 
	\centering 
	\includegraphics[width=0.95\textwidth]{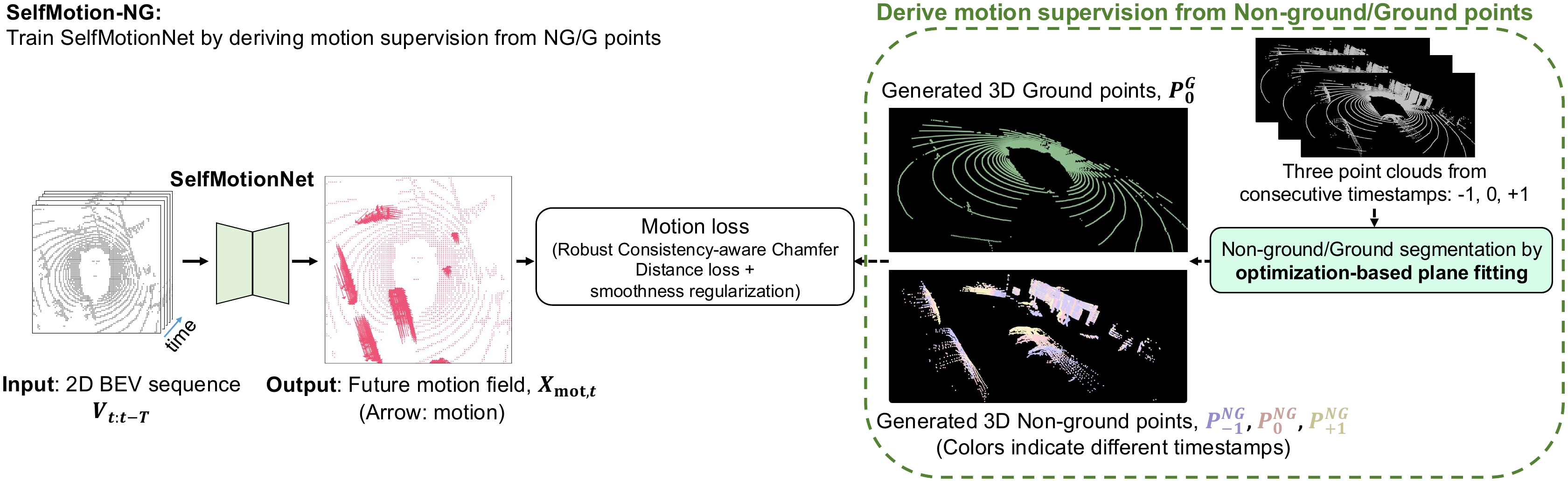}
	\vspace{-2mm}
	\caption{Overview of \textbf{SelfMotion-NG}, a self-supervised motion prediction approach guided by \textbf{non-ground/ground cues}.
		In {SelfMotion-NG}, we train a motion prediction network, \textbf{SelfMotionNet}, to predict future motion field~${\bm X_{{\rm  mot},t}}$ from a sequence of synchronized BEV maps $\bm V_{t:t-T}$.
		To derive motion supervision, we use an \textbf{optimization-based plane fitting} to generate non-ground/ground points, and apply a Robust Consistency-aware Chamfer Distance loss to these points for self-supervised motion learning.}
	\label{fig_self_pipeline}
\end{figure*}

\subsubsection{Implementation of WeakMotion-NG}

In WeakMotion-NG, we adopt the same motion prediction network, WeakMotionNet, as in \mbox{WeakMotion-FB}.

For the training of the motion prediction head in WeakMotionNet, we use the ground segmentation method presented in Sec.~\ref{chap_ground_seg} to generate non-ground points from the past $(t\text{-}1)$, current $(t)$ and future $(t\text{+}1)$ timestamps, denoted as  ${\bm P_{t-1}^{\rm NG}}$, ${\bm P_{t}^{\rm NG}}$, and ${\bm P_{t+1}^{\rm NG}}$, respectively.
 The generated ground points at the current~$(t)$  timestamp are denoted as ${\bm P_{t}^{\rm G}}$.

Following Eq.~(\ref{eq_BEV_to_point}), we map the predicted BEV motion ${\bm X_{{\rm  mot},t}}$ to the points at the current timestamp to obtain the point-wise motion prediction ${\bm F_t}$ for ${\bm P_{t}}$.
The motion loss of the motion prediction head is written as:
\begin{equation}\small\label{Self_loss}
\begin{aligned}
{\cal L}_{\rm Self,mot}  = &{\cal L}_{RCCD}({\bm P^{\rm NG}_{t\text{-}1}}, {\bm P^{\rm NG}_{t}}, {\bm P^{\rm NG}_{t\text{+}1}}, {\bm F_{t}^{\rm NG}}) + {\cal L}_{\rm smooth}({\bm F_t^{\rm NG}})  \\
& +  \varphi_{g} {\cal L}_{\rm mot,S} ({\bm F_t^{\rm G}}), 
\end{aligned}
\end{equation}
where the first term is the RCCD loss (Eq.~(\ref{CCD})) for the predicted motion of the non-ground points, ${\bm F_t^{\rm NG}}$;  the second term is the smoothness regularization  (Eq.~(\ref{eq.smooth}))  for ${\bm F_t^{\rm NG}}$; and the third term, defined in Eq.~(\ref{eq.mot_S}), is for the predicted motion of the ground points, ${\bm F_t^{\rm G}}$.
Here, $\varphi_{g}$ denotes the weight coefficient for the motion loss of the ground points.

For the training of auxiliary FG/BG segmentation head in WeakMotionNet, we use the same classification loss, $ {\cal L}_{\rm cls} $, (Eq.~(\ref{cls})) and follow the same strategy as in WeakMotion-FB.
The total loss in WeakMotion-NG can be written as: 
\begin{equation}
{\cal L}_{\rm wng} = \omega \cdot \beta_1 {\cal L}_{\rm cls} + \beta_2 {\cal L}_{\rm Self,mot},
\end{equation} 
where $\beta_1$, $\beta_2$ are the weight coefficients for different loss terms, while $\omega$ denotes the weight coefficient for different training samples.
Specifically, we set $\beta_1$ to $1$ and $\beta_2$ to $5$.
Since we only partially annotate  one frame per sequence, $\omega$ is $1$ if this frame is partially annotated; otherwise, 0.

\subsection{Motion Prediction with Self-supervision}\label{sec_self}
By removing the auxiliary FG/BG segmentation, we further extend  WeakMotion-NG proposed in Sec.~\ref{sec_weaker} into a self-supervised motion prediction approach, \textbf{SelfMotion-NG}, thereby eliminating the need for FG/BG annotations.

As presented in Fig.~\ref{fig_self_pipeline}, the self-supervised motion prediction network, \textbf{SelfMotionNet}, takes a sequence of synchronized BEV maps as input and predicts future motion displacement for each cell. 
Specifically, SelfMotionNet consists of a backbone network and a motion prediction head, both sharing the same architecture as those in WeakMotionNet.
In the absence of motion data,  we adopt the same ground segmentation method described in Sec.~\ref{chap_ground_seg} to extract ground and non-ground points.
We then train SelfMotionNet using the same motion loss, ${\cal L}_{\rm Self,mot}$, defined in Eq.~(\ref{Self_loss}), to enable self-supervised motion prediction.

\section{Experiments}

In this section, we first present the experimental setup in  Sec.~\ref{Es}.
Then, we compare our models with state-of-the-art  supervised, semi-supervised, and self-supervised motion prediction methods in Sec.~\ref{SOTA}.
Finally, we conduct ablation studies to analyze the effectiveness of each component in Sec.~\ref{AS}.

\subsection{Experimental setup}\label{Es}
\subsubsection{Dataset}
The main experiments are conducted on nuScenes~\cite{caesar2020nuscenes}, a large-scale autonomous driving dataset.
Following previous works~\cite{wu2020motionnet,luo2021self,wang2022sti}, we adopt 500 scenes for training, 100 for validation, and 250 for testing.
For each scene, we utilize the LiDAR point clouds as input. 
During the training stage, we use the officially annotated foreground and background labels from nuScenes  as weak supervision. 
Specifically, nuScenes defines 32 semantic classes, among which 23 (e.g., vehicles, pedestrians, cyclists) are officially labeled as foreground and the remaining 9 (e.g., buildings, vegetation) as background.
In the validation and testing stage, we generate motion data from  detection and tracking annotations provided by nuScenes as ground truth for evaluation.

Additionally, we apply our approaches to Waymo Open Dataset~\cite{sun2020scalability}.
Compared to nuScenes, which uses a 32-beam LiDAR sensor, the Waymo dataset employs 64-beam LiDAR sensors, resulting in  denser and higher-quality point clouds.  
Specifically, we extract 14,351 samples from training set for training and 3,634 samples from validation set for testing. 
The motion ground truth for the Waymo dataset is also derived from object detection and tracking annotations. The bounding boxes are classified into four categories: vehicle, pedestrian, cyclist, and sign. 
Accordingly, we treat vehicles, pedestrians, and cyclists as foreground, and consider all remaining points in the scene as background.

\begin{table*}[t!]
	\caption{Evaluation results of motion prediction on nuScenes test set. Full., Semi., Self., Weak., refer to fully-supervised, semi-supervised, self-supervised, and  weakly supervised training respectively.
		With fully (100\%) or partially (1\%, 0.1\%, 0.01\%) annotated FG/BG masks as weak supervision, our models outperform the self-supervised models by a large margin, and perform on par with some supervised ones, indicating that our weakly supervised approaches achieve a good compromise between annotation effort and performance.  
		Additionally, without using any annotations, our self-supervised approach achieves  state-of-the-art performance in self-supervised motion prediction.
		\protect \tikz  \protect \draw[tabblue,fill=tabblue] (0,0) circle (.7ex); means partially annotating 0.1\% of points in one frame of a 10-frame sequence (the overall ratio is about 0.01\%). 
		\protect \tikz  \protect \draw[taborange,fill=taborange] (-.7ex,-.7ex)  rectangle (.7ex,.7ex); means partially annotating 1\% or 0.1\% of points per frame.
		\protect \tikz  \protect \draw[tabgreen,fill=tabgreen] (0,.7ex) -- (-.7ex,-.7ex) -- (.7ex,-.7ex) -- cycle; means fully annotating every point per frame.}
	\vspace{-5mm}
	\label{tab_nuscenes}
	\begin{center}
		\setlength{\tabcolsep}{4.5pt}
		\renewcommand\arraystretch{1.1}
		\resizebox{1.95\columnwidth}{!}{
			\centering
			\begin{tabular}{ l |c| c  | cc | cc | cc }
				\Xhline{1.4pt}
				\multirow{2}{*}{Method} & \multirow{2}{*}{Supervision} & \multirow{2}{*}{Modality} &  \multicolumn{2}{c|}{Static} & \multicolumn{2}{c|}{Speed $\leq$ 5m/s} &  \multicolumn{2}{c}{Speed $>$ 5m/s} \\  \cline{4-9}
				& & & Mean~$\downarrow$ & Median~$\downarrow$ & Mean~$\downarrow$ & Median~$\downarrow$ & Mean~$\downarrow$ & Median~$\downarrow$\\
				\Xhline{1.4pt}
				FlowNet3D \cite{liu2019flownet3d} & Full. & LiDAR& 0.0410 & 0 & 0.8183 & 0.1782 & 8.5261 & 8.0230\\
				HPLFlowNet \cite{gu2019hplflownet} & Full.  & LiDAR &  0.0041 & 0.0002 & 0.4458 & 0.0960 & 4.3206 & 2.4881 \\
				PointRCNN \cite{shi2019pointrcnn} & Full.  & LiDAR & 0.0204 & 0 & 0.5514 & 0.1627 & 3.9888 & 1.6252 \\
				LSTM-ED \cite{schreiber2019long}&  Full. & LiDAR & 0.0358 & 0 & 0.3551 & 0.1044 & 1.5885 & 1.0003 \\
				PillarMotion~\cite{luo2021self} & Full. & LiDAR+Image & 0.0245 & 0 & 0.2286 & 0.0930 & 0.7784 & 0.4685\\
				MotionNet~\cite{wu2020motionnet}  & Full. & LiDAR & 0.0201 & 0 & 0.2292 & 0.0952 & 0.9454 & 0.6180\\
				BE-STI~\cite{wang2022sti}  &  Full. & LiDAR & 0.0220 & 0 & 0.2115 & 0.0929 & 0.7511 & 0.5413\\
				SSMP~\cite{wang2024semi}  &  Semi. (1\% motion ground truth) & LiDAR & 0.0153 & 0 & 0.3497 & 0.1020 & 1.9407 & 1.2173\\
				\Xhline{1.4pt}
				PillarMotion~\cite{luo2021self} & Self. &  LiDAR+Image & 0.1620 &  0.0010 &  0.6972&  0.1758& 3.5504 & 2.0844\\
				ContrastMotion~\cite{jia2023contrastmotion}  & Self. &  LiDAR & 0.0829 &  0 &  0.4522 & \underline{0.0959} & 3.5266 & 1.3233 \\
				RigidFlow++~\cite{li2024self}  & Self. &  LiDAR & 0.0580 &  0 &  \underline{0.3097} & {0.1001} & 2.4937 & 1.2662 \\
				SSMotion~\cite{fang2024self}  & Self. &  LiDAR & 0.0514 & 0 &0.4212 & 0.1073 &  \bf 2.0766 & 1.3226 \\
				SelfMotion~\cite{wang2024self}  & Self. &  LiDAR & {\bf 0.0419} &  0 &  0.3213 & 0.1061 & 2.2943 &\underline{1.0508}\\
				Ours: SelfMotion-NG (0\%)& Self. & LiDAR&    \underline{0.0439}   & {\bf 0} &  \bf 0.2656    & \bf 0.0943     & \underline{2.1838} &   \bf 1.0236    \\
				\Xhline{1.4pt}
				\rowcolor{gray}
				Ours: WeakMotion-NG (0.01\%)& \tikz\draw[tabblue,fill=tabblue] (0,0) circle (.7ex); Weak. (0.01\% FG/BG masks) & LiDAR& 0.0374  & 0 & 0.2466   &  \underline{ 0.0938 }  & 1.6421    &  0.8246   \\
				WeakMotion~\cite{li2023weakly} (0.1\%) &  { \protect \tikz  \protect \draw[taborange,fill=taborange] (-.7ex,-.7ex)  rectangle (.7ex,.7ex);  Weak.   \, (0.1\% FG/BG masks) }  & LiDAR&  0.0426  & 0 & 0.4009  & 0.1195  & 2.1342  & 1.2061 \\
				Ours: WeakMotion-FB (0.1\%)&  { \protect \tikz  \protect \draw[taborange,fill=taborange] (-.7ex,-.7ex)  rectangle (.7ex,.7ex);  Weak.   \, (0.1\% FG/BG masks) } & LiDAR& 0.0259  & 0 & 0.2472   & 0.0942  & 1.4037  & 0.8252 \\
				\rowcolor{gray}
				WeakMotion~\cite{li2023weakly} (1\%)  & { \protect \tikz  \protect \draw[taborange,fill=taborange] (-.7ex,-.7ex)  rectangle (.7ex,.7ex);  Weak. \,	 \quad (1\% FG/BG masks)}    & LiDAR& 0.0558  & 0 & 0.4337  & 0.1305   & {1.7823} & {1.0887} \\
				\rowcolor{gray}
				Ours: WeakMotion-FB (1\%)&  { \protect \tikz  \protect \draw[taborange,fill=taborange] (-.7ex,-.7ex)  rectangle (.7ex,.7ex);  Weak.  \, \quad (1\% FG/BG masks)}  & LiDAR& \underline{0.0231} & 0 & \underline{0.2390} & \underline{0.0938}  & \underline{1.2108} & \underline{0.7555}  \\
				WeakMotion~\cite{li2023weakly} (100\%)& {\protect \tikz  \protect \draw[tabgreen,fill=tabgreen] (0,.7ex) -- (-.7ex,-.7ex) -- (.7ex,-.7ex) -- cycle; Weak. \, (100\% FG/BG masks)} & LiDAR&  0.0243 &  0 &  0.3316 &  0.1201  &  1.6422 &  1.0319 \\
				Ours: WeakMotion-FB (100\%)& {\protect \tikz  \protect \draw[tabgreen,fill=tabgreen] (0,.7ex) -- (-.7ex,-.7ex) -- (.7ex,-.7ex) -- cycle; Weak. \, (100\% FG/BG masks)}   & LiDAR& {\bf{0.0224}} &  {\bf 0} & {\bf 0.2366} & \bf{0.0931}  &{\bf 1.1974} &{\bf 0.7456} \\
				\Xhline{1.4pt}
			\end{tabular}
		} 
	\end{center}
\end{table*}

\begin{table}
	\caption{ Results of FG/BG segmentation on nuScenes test set. }
	\vspace{-5mm}
	\label{tab_nuscenes_class}
	\begin{center}
		\setlength{\tabcolsep}{3.0pt}
		\resizebox{0.95\columnwidth}{!}{
			\centering
			\begin{tabular}{l|@{\hskip 0.1cm}c@{\hskip 0.1cm}|@{\hskip 0.1cm}c@{\hskip 0.1cm}|@{\hskip 0.1cm}c@{\hskip 0.1cm}}
				\Xhline{1.2pt}
				Method& { BG Acc.~$\uparrow$} &{ FG Acc.~$\uparrow$}   & {Overall Acc.~$\uparrow$}  \\
				\Xhline{1.2pt}
				Ours: WeakMotion-NG (0.01\%) & 95.6\% & 75.7\%  &  94.4\%  \\
				Ours: WeakMotion-FB (0.1\%)&    95.7\% &  	87.9\%  & 95.2\% \\
				Ours: WeakMotion-FB (1.0\%)&    95.6\% & 92.0\%  &  95.4\%  \\
				Ours: WeakMotion-FB (100\%)&    93.7\% &  95.1\%  &  93.8\%    \\
				\Xhline{1.2pt}
			\end{tabular}
		}
	\end{center}
\end{table}

\begin{table}
	\caption{Motion prediction results on Waymo Open Dataset. 
		Our weakly and self-supervised approaches outperform self-supervised SelfMotion~\cite{wang2024self} across all evaluation metrics. 
		\protect \tikz \protect \draw[tabblue,fill=tabblue] (0,0) circle (.7ex); means partially annotating 0.01\% of points in one frame of a 10-frame sequence (the overall ratio is about 0.001\%). 
		\protect \tikz  \protect \draw[taborange,fill=taborange] (-.7ex,-.7ex)  rectangle (.7ex,.7ex); means partially annotating 1\% or 0.1\% of points per frame.
		\protect \tikz  \protect \draw[tabgreen,fill=tabgreen] (0,.7ex) -- (-.7ex,-.7ex) -- (.7ex,-.7ex) -- cycle; means fully annotating every point per frame.}
	\vspace{-7mm}
	\label{tab_waymo}
	\begin{center}
		\setlength{\tabcolsep}{3.0pt}
		\renewcommand\arraystretch{1.1}
		\resizebox{1.0\columnwidth}{!}{
			\centering
			\begin{tabular}{l|c|c@{\hskip 0.05cm}c@{\hskip 0.05cm}c}
				\Xhline{1.2pt}
				{Method}& Supervision&{Static$\downarrow$} &{Sp.$\leq$5m/s$\downarrow$}  & { Sp.$>$5m/s$\downarrow$}  \\
				\Xhline{1.2pt}
				MotionNet~\cite{wu2020motionnet}  & Full. & 	0.0263   &  0.2620   & 0.9493   \\
				\Xhline{1.2pt}
				SelfMotion~\cite{wang2024self} & Self.  &   0.0515 &  0.4440  & 2.3692\\
				Ours: SelfMotion-NG (0\%)& Self.  &   \bf 0.0418  &  \bf 0.3394  &  \bf 1.8110   \\
				\Xhline{1.2pt}
				\rowcolor{gray}
				Ours: WeakMotion-NG (0.001\%)& \tikz\draw[tabblue,fill=tabblue] (0,0) circle (.7ex); Weak. (0.001\% FG/BG)  &    0.0402  &  0.3178  & 1.5540    \\
				WeakMotion~\cite{li2023weakly} (0.1\%)&  { \protect \tikz  \protect \draw[taborange,fill=taborange] (-.7ex,-.7ex)  rectangle (.7ex,.7ex);  Weak.   \, (0.1\% FG/BG) }  &  {0.0297}    &   0.3581   & {1.6362}    \\
				Ours: WeakMotion-FB (0.1\%)&  { \protect \tikz  \protect \draw[taborange,fill=taborange] (-.7ex,-.7ex)  rectangle (.7ex,.7ex);  Weak.   \, (0.1\% FG/BG) }  & 0.0295 &0.3215 & 1.2830     \\
				\rowcolor{gray}
				WeakMotion~\cite{li2023weakly} (1\%)&  { \protect \tikz  \protect \draw[taborange,fill=taborange] (-.7ex,-.7ex)  rectangle (.7ex,.7ex);  Weak. \,	 \quad (1\% FG/BG)} &  0.0334   &   {0.3458}    &    {1.5655}   \\
				\rowcolor{gray}
				Ours: WeakMotion-FB (1.0\%)&{ \protect \tikz  \protect \draw[taborange,fill=taborange] (-.7ex,-.7ex)  rectangle (.7ex,.7ex);  Weak. \,	 \quad (1\% FG/BG)}  &0.0291 	& \underline{0.3046} & \bf 1.2136    \\
				WeakMotion~\cite{li2023weakly} (100\%)& {\protect \tikz  \protect \draw[tabgreen,fill=tabgreen] (0,.7ex) -- (-.7ex,-.7ex) -- (.7ex,-.7ex) -- cycle; Weak. \, (100\% FG/BG)} &    \underline{0.0219}  &    0.3385   &   {1.6576}  \\
				Ours: WeakMotion-FB (100\%)&  {\protect \tikz  \protect \draw[tabgreen,fill=tabgreen] (0,.7ex) -- (-.7ex,-.7ex) -- (.7ex,-.7ex) -- cycle; Weak. \, (100\% FG/BG)} &   \bf 0.0205  &	\bf 0.2981 	& \underline{1.2342}   \\
				\Xhline{1.2pt}
			\end{tabular}
		}
	\end{center}
\end{table}

\begin{figure*}[htb]
	\centering 
	\includegraphics[width=0.97\textwidth]{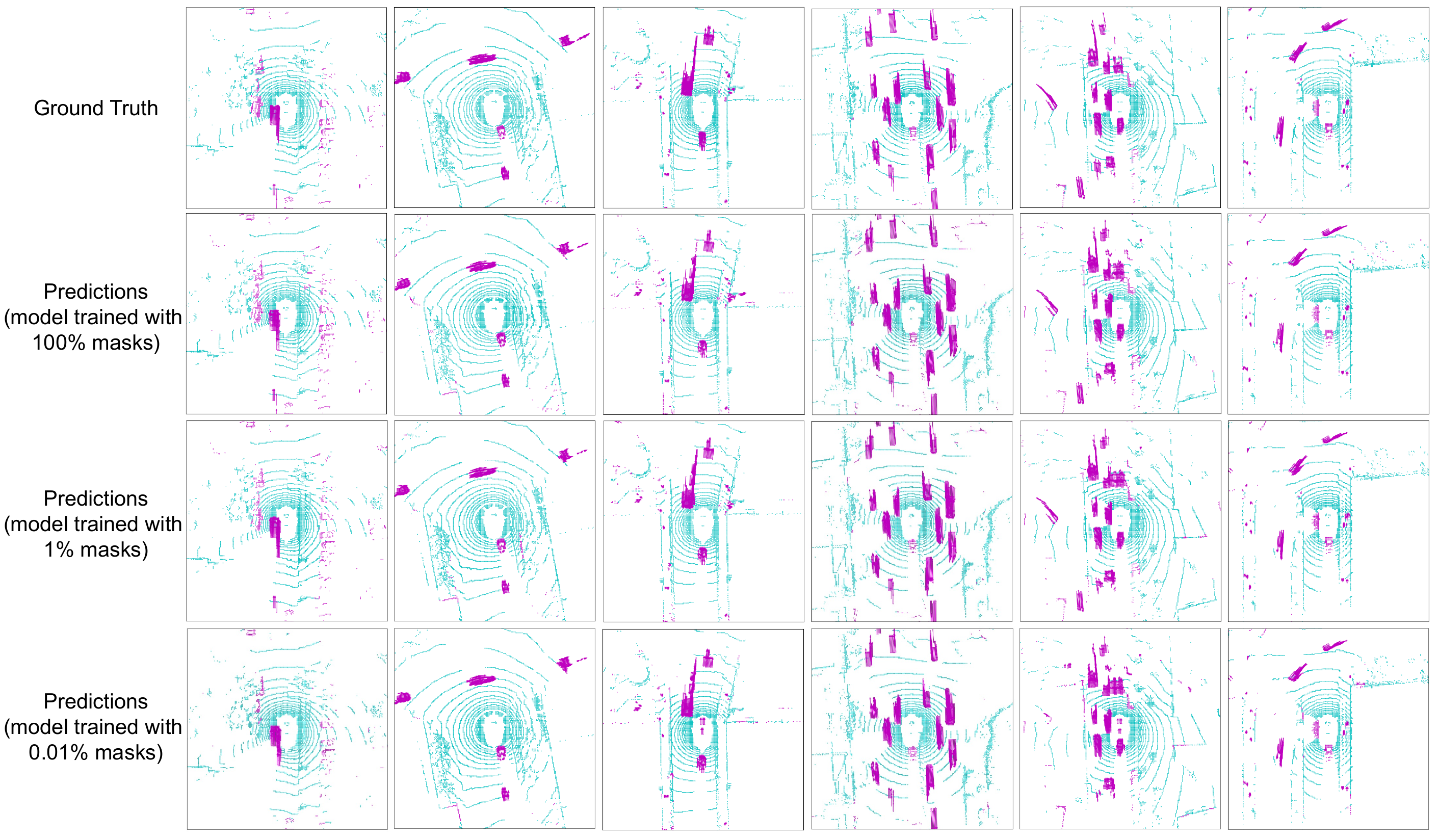}
	\vspace{-4mm}
	\caption{Qualitative results of motion prediction and foreground/background segmentation predicted by our weakly supervised models \mbox{(WeakMotionNet)} on nuScenes. We show motion with an arrow attached to each cell and represent different categories with different colors.
		{\textcolor[RGB]{255,0,255}{\textbf{Purple}}}: Foreground;  	{\textcolor[RGB]{0,228,238}{\textbf{Cyan}}}: Background.
	Although trained only with foreground/background masks as weak supervision, our models can effectively generate motion and segmentation predictions. }
	\label{fig_results}	
\end{figure*}

\begin{figure*}[htb]
	\centering 
	\includegraphics[width=0.97\textwidth]{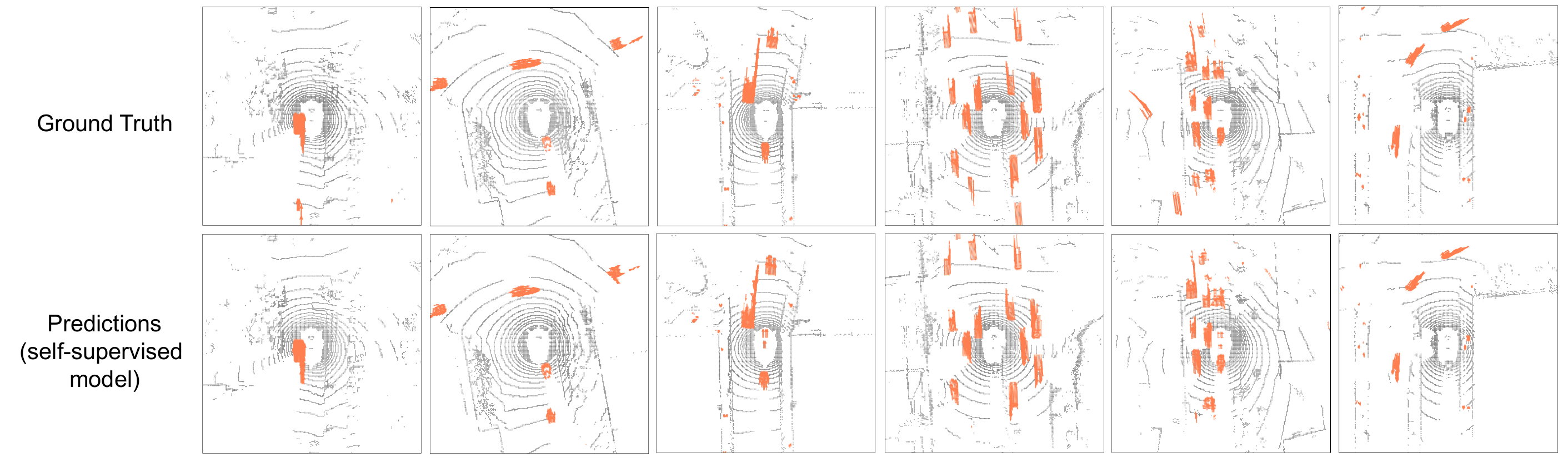}
	\vspace{-4mm}
	\caption{
		Qualitative results of motion prediction and static/moving segmentation predicted by our self-supervised model \mbox{(SelfMotionNet)} on nuScenes.
		We show motion with an arrow attached to each cell  and represent different motion states with different colors.
		{\textcolor[RGB]{237,125,49}{\textbf{Orange}}}: Moving objects;  	{\color{Gray}{\textbf{Gray}}}: Static objects. 
		Trained without any annotations, our self-supervised model can still achieve reasonable motion prediction.}
	\label{fig_results_self}	
\end{figure*}

\subsubsection{Implementation details}
\noindent\textbf{Experiments on nuScenes dataset.}\quad 
Following the same data pre-processing settings in~\cite{wu2020motionnet,wang2022sti}, we  crop each input point cloud in the range of  $[-32, 32]\times[-32, 32] \times [-3, 2]$ meters and set the voxel size to be $(0.25, 0.25, 0.4) {\rm m}$ for nuScenes.  

For fully and per-frame partially (1\%, 0.1\%) annotated FG/BG masks, we apply  \mbox{\textbf{WeakMotion-FB}}  for training.
Specifically, we train the  FG/BG segmentation network, \textbf{PreSegNet}, using partially annotated FG/BG masks as weak supervision for 40 epochs.
For the classification loss of PreSegNet, we set the weight $\alpha_{\tau}(i)$ to  $0.005$ if the ground truth label of point $i$ is background (BG); otherwise, $1$.
Additionally, we set the batchsize to 32 and use Adam~\cite{kingma2014adam} with an initial learning rate of 0.0005, decayed by 0.5 after every 10 epochs.

Subsequently, we train the motion prediction network, \mbox{\textbf{WeakMotionNet}}, with a sequence of point clouds as input.
For fair comparisons with~\cite{wu2020motionnet,luo2021self,wang2022sti}, we set the sequence length to 5.
Each input sequence contains 1 current frame and 4 past frames, with a time span of 0.2 seconds between consecutive frames.	
Following~\cite{luo2021self}, the WeakMotionNet is designed to output the displacement for the next 0.5s as the predicted motion.
Correspondingly, in self-supervised motion learning, the past frame and the future frame  are the point clouds of the past 0.5s and the next 0.5s, respectively.
And the trained PreSegNet  will generate FG and BG points for the past, current, and future frames.
Note that when using fully annotated masks as supervision, we directly use the ground truth FG and BG points for self-supervised motion learning.
We train the WeakMotionNet for 80 epochs with an initial learning rate of 0.0005, and we decay it by 0.5 after every 10 epochs.
We set the background weight $\varphi_{bg}$ to 0.005, the batchsize to 16, and use Adam as optimizer.

When partially annotating 0.1\% of points in one frame of a 10-frame sequence (the overall ratio is about 0.01\%), we adopt \mbox{\textbf{WeakMotion-NG}}  to train the motion prediction network, \mbox{\textbf{WeakMotionNet}}, following the same training strategy used in  WeakMotion-FB.
Specifically, we set the ground weight $\varphi_{g}$ to 0.005.

For self-supervised motion prediction, we adopt \textbf{SelfMotion-NG}  to train the motion prediction network, \mbox{\textbf{SelfMotionNet}}, using the same training strategy as that employed for WeakMotionNet in WeakMotion-NG.
In this case, we set the ground weight $\varphi_{g}$ to 0.02.\\

\noindent\textbf{Experiments on Waymo Open Dataset.}\quad 
For Waymo, we set range to $[-32, 32]\times[-32, 32] \times [-1, 4]$ and  voxel size to  $(0.25, 0.25, 0.4) {\rm m}$. 
The training strategies for Waymo are the same as those for nuScenes, with some adjustments to the weight coefficients and the batch size.

For \mbox{WeakMotion-FB}, we set the background weight $\varphi_{bg}$ to $0.02$, the batch size to 12, and the weight $\alpha_{\tau}(i)$ to $0.02$ for background points.
For \mbox{WeakMotion-NG}, we set the ground weight $\varphi_{g}$ to $0.005$, the batch size to 8, and the weight $\alpha_{\tau}(i)$ to $0.005$ for background points. 
In particular, since the number of points per frame in Waymo is larger than in nuScenes, we can annotate fewer points per frame when using \mbox{WeakMotion-NG} in Waymo. Specifically, we partially annotate 0.01\% of points in one frame of a 10-frame sequence, resulting in an overall annotation ratio of approximately 0.001\%.
For  SelfMotion-NG, we  set the ground weight $\varphi_{g}$ to $0.005$ and the batch size to 8.
All of our methods are implemented  in PyTorch~\cite{paszke2019pytorch}.

\subsubsection{Evaluation metrics}
For the motion prediction, following previous works~\cite{wu2020motionnet,luo2021self,wang2022sti},  we divide non-empty cells into three groups: static, slow ($\leq 5{\rm m/s}$), fast ($\geq 5{\rm m/s}$) and evaluate the mean and median errors  on each group.
Errors are measured by  $L_2$ distances between the predicted displacements and the ground truth displacements for the next 1s. 
The outputs of our WeakMotionNet and SelfMotionNet are the displacements for the next 0.5s.
Therefore, we assume that the speed is constant within a short time windows and linearly extrapolate the outputs to the next 1s for evaluation.
For the FG/BG segmentation, we measure the accuracy of each category (Acc.), i.e., the proportion of correctly predicted cells in that category, and overall classification accuracy (Overall Acc.), i.e., the average accuracy over all non-empty cells.

\begin{table*}
	\caption{Ablation study of Robust Consistency-aware Chamfer Distance (RCCD) loss on WeakMotion-FB  with 1\% FG/BG annotation ratio. WL means the Welsch-Leclerc penalty; GM means the Geman-McClure penalty; Smooth Reg. means spatial smoothness regularization. FG/BG Seg. means auxiliary FG/BG segmentation head and its classification loss.}
	\vspace{-6mm}
	\label{tab_CCD}
	\begin{center}
		\setlength{\tabcolsep}{3.0pt}
		\renewcommand\arraystretch{1.1}
		\resizebox{2.0\columnwidth}{!}{
			\centering
			\begin{tabular}{l|c@{\hskip 0.1cm}c@{\hskip 0.1cm}c@{\hskip 0.1cm}c@{\hskip 0.05cm}|c@{\hskip 0.1cm}c@{\hskip 0.05cm}|@{\hskip 0.05cm}c@{\hskip 0.05cm}|@{\hskip 0.05cm}c@{\hskip 0.05cm}|@{\hskip 0.05cm}c@{\hskip 0.05cm}|l@{\hskip 0.1cm}l@{\hskip 0.1cm}l@{\hskip 0.1cm}}
				\Xhline{1.6pt}
				\multirow{2}{*}{Loss function in WeakMotionNet} & \multicolumn{4}{c|}{Penalty} & Future & Past & Confidence& Smooth &  FG/BG   &{Static} &{Sp.$\leq$5m/s}  & { Sp.$>$5m/s}  \\
				
				& $L_2$  & $L_1$ & WL & GM  & Frame &  Frame & Reweight& Reg. & Seg.&  \multicolumn{3}{c}{Mean Error~$\downarrow$}   \\
				\Xhline{1.6pt}
				(a) Chamfer-$L_2$ (\bf Baseline) &   \checkmark & && & \checkmark & & &	&&	0.524 & 0.824 &2.274 \\
				\hline
				(b) Chamfer-$L_1$   &  & \checkmark  &&  & \checkmark &  & &  &	&   0.205$\,$\green{61} &  0.429$\,$\green{48} &  1.775$\,$\green{22} \\
				(c) Chamfer-WL   &  &   & \checkmark &  & \checkmark &  & & 	&	& 0.087$\,$\green{83} & 0.444$\,$\green{46} &  1.741$\,$\green{23}  \\
				(d) Chamfer-GM   &  &   &  & \checkmark & \checkmark &  & &  &	& 0.077$\,$\green{85}  &	0.406$\,$\green{51} & 1.662$\,$\green{27}  \\
				\hline
				(e) Multi-frame Chamfer-GM &  &   &&\checkmark & \checkmark &  \checkmark & & 	& &  0.083$\,$\green{84}  & 0.413$\,$\green{50}  &   1.464$\,$\green{36}  \\
				(f) Consistency-aware Chamfer-$L_1$  ({\bf CCD loss} in~\cite{li2023weakly})  &  &  \checkmark &&  & \checkmark &  \checkmark& \checkmark& 			&  &  0.103$\,$\green{80}	& 0.389$\,$\green{53}  &  1.571$\,$\green{31}  \\
				(g) Robust Consistency-aware Chamfer  ({\bf Our RCCD loss}) &  &  && \checkmark & \checkmark &  \checkmark& \checkmark& 	& 	& 0.060$\,$\green{89}  &  0.396$\,$\green{52}  & 1.574$\,$\green{31}  \\
				(h) Robust Consistency-aware Chamfer  + Smooth &  &   &&\checkmark & \checkmark &  \checkmark& \checkmark& 		\checkmark	&  & 0.039$\,$\green{93}  & \bf0.236$\,$\green{\bf71}  & \bf1.134$\,$\green{\bf50} \\
				\hline
				(i) Robust Consistency-aware Chamfer + Smooth + Seg.&  &   && \checkmark & \checkmark &  \checkmark& \checkmark& \checkmark & \checkmark	 & \bf 0.023$\,$\green{\bf96}  &  \bf 0.239$\,$\green{\bf71}  &  1.211$\,$\green{47}   \\
				\Xhline{1.6pt}
			\end{tabular}
		}
	\end{center}
\end{table*}

\begin{table*}
	\caption{Effectiveness of our weakly supervised training approaches, WeakMotion-FB and WeakMotion-NG, under different FG/BG annotation ratios (1\% and 0.01\%). Slow means speed $\leq 5{\rm m/s}$. Fast means  speed $\geq 5{\rm m/s}$. We report the mean error on each speed group and the accuracy of each category for evaluation. }
	\vspace{-8mm}
	\label{tab_Stage}
	\begin{center}
		\setlength{\tabcolsep}{3.0pt}
		\renewcommand\arraystretch{1.2}
		\resizebox{2.05\columnwidth}{!}{
			\centering
			\begin{tabular}{l@{\hskip 0.05cm}|c@{\hskip 0.1cm}c@{\hskip 0.1cm}c|c@{\hskip 0.1cm}c@{\hskip 0.05cm}c@{\hskip 0.05cm}|c@{\hskip 0.1cm}c@{\hskip 0.1cm}c|c@{\hskip 0.1cm}c@{\hskip 0.05cm}c}
				\Xhline{1.2pt}
				\multirow{2}{*}{Weakly supervised method} & \multicolumn{6}{c|}{1\% FG/BG masks} & \multicolumn{6}{c}{0.01\% FG/BG masks}  \\
				\cline{2-13}
				& {Static$\downarrow$} &{Slow$\downarrow$}  & {Fast$\downarrow$} & {BG$\uparrow$} &{FG$\uparrow$}  & {Overall$\uparrow$}  & {Static$\downarrow$} &{Slow$\downarrow$}  & {Fast$\downarrow$} & {BG$\uparrow$} &{FG$\uparrow$}  & {Overall$\uparrow$}  \\
				\Xhline{1.2pt}
				Motion supervision from raw points &  0.086 & 0.502 & 6.093 & 84\% & 97\% & 85\%  &  0.126 &0.496 & 6.363 & 83\% & 87\% & 83\% \\
				Motion supervision from predicted non-ground points ({\bf WeakMotion-NG}) & \bf 0.022 &0.249 & 1.533 & 93\% &  94\% & 93\%  &   \bf 0.037 & \bf 0.247 & \bf 1.642 &	96\% & 76\% & 94\% \\
				Motion supervision from predicted foreground points ({\bf WeakMotion-FB}) &  \bf 0.023 & \bf 0.239 & \bf 1.211 & 96\% & 92\% & 95\%  &  0.046 &0.261 & 1.810 & 95\% & 78\% & 94\%  \\
				\Xhline{1.2pt}
			\end{tabular}
		}
	\end{center}
\end{table*}

\subsection{Comparison with State-of-the-Art Methods}\label{SOTA}

\subsubsection{Results on  nuScenes dataset}
In Table~\ref{tab_nuscenes}, we compare  our weakly and self-supervised approaches with various state-of-the-art motion prediction methods on nuScenes~\cite{caesar2020nuscenes}.

As shown in Table~\ref{tab_nuscenes}, our four weakly supervised models outperform the competing self-supervised models on all evaluation metrics and perform better than the supervised FlowNet3D~\cite{liu2019flownet3d}, HPLFlowNet~\cite{gu2019hplflownet}, and PointRCNN~\cite{shi2019pointrcnn} on both slow and fast speed groups.
In particular, our weakly supervised model trained with only 0.01\% annotated FG/BG masks surpasses the fully supervised scene flow models, FlowNet3D and HPLFlowNet, by about {\bf 60\%}  on the fast speed group.
Additionally, it outperforms the semi-supervised SSMP~\cite{wang2024semi} trained with 1\% motion annotations by about {\bf 8\%} on the slow speed group and about {\bf 15\%} on the fast speed group, despite using only 0.01\% FG/BG masks as weak supervision.
These comparisons demonstrate that our weakly supervised approaches strike a good balance between annotation effort and performance, and further narrow the gap with fully supervised methods.

Notably, as shown in Table~\ref{tab_nuscenes}, under different annotation ratios (0.1\%, 1\%, and 100\%), our WeakMotion-FB approach outperforms the original version, WeakMotion~\cite{li2023weakly}, by a significant margin on all evaluation metrics. This indicates that our proposed RCCD loss and its smoothness regularization achieve substantial improvements over the original CCD loss in WeakMotion.
Additionally, our proposed WeakMotion-NG approach reduces the FG/BG annotation ratio to 0.01\%, while still delivering better results than WeakMotion~\cite{li2023weakly} trained with 1\% FG/BG annotations.
Furthermore, without using any annotations, our  self-supervised approach, SelfMotion-NG, achieves superior performance  in self-supervised motion prediction.

The performance of the FG/BG segmentation head of WeakMotionNet is shown in Table~\ref{tab_nuscenes_class}.
Despite being trained with a tiny fraction of annotated masks (0.01\%, 0.1\%, and 1\%), our models can distinguish foreground and background with high overall  accuracy (about 94\%).
The qualitative results of our weakly supervised models and our self-supervised model are depicted in Fig.~\ref{fig_results} and Fig.~\ref{fig_results_self}, respectively.
With a sequence of BEV maps as input, our WeakMotionNet and SelfMotionNet take 16ms for inference in a  RTX A5000 GPU.

\subsubsection{Results on  Waymo Open Dataset}
For further evaluation, we also apply our weakly and self-supervised approaches to Waymo~\cite{sun2020scalability}.
As shown in Table~\ref{tab_waymo}, our weakly and self-supervised models significantly outperform the self-supervised SelfMotion~\cite{wang2024self} across all evaluation metrics. 
Under different annotation ratios (0.1\%, 1\%, and 100\%), our weakly supervised approach, WeakMotion-FB, surpasses the original WeakMotion~\cite{li2023weakly} in both slow and fast speed groups, demonstrating the effectiveness of  our proposed RCCD loss and smoothness regularization over the original CCD loss used in WeakMotion. 
Furthermore, our proposed WeakMotion-NG approach achieves comparable performance to WeakMotion~\cite{li2023weakly} trained with 1\% FG/BG masks, despite using only 0.001\% FG/BG masks.

\subsection{Ablation Studies}\label{AS}
In this subsection, we evaluate the effectiveness of our weakly and self-supervised approaches on nuScenes.
Specifically, we first analyze the contributions of our proposed Robust Consistency-aware Chamfer Distance loss, and subsequently validate the effectiveness of our weakly and self-supervised motion prediction approaches.
Finally, we conduct a visualization analysis to provide better insights into our approaches.

\subsubsection{Ablation study for Robust Consistency-aware Chamfer Distance loss.}

For robust self-supervised motion learning, we design a Robust Consistency-aware Chamfer Distance loss with a robust penalty function as distance metric, multi-frame point clouds for supervision, and multi-frame consistency for reweighting.
To analyze the effectiveness of the proposed RCCD loss, we conduct an ablation study on WeakMotion-FB using 1\% FG/BG masks as weak supervision.

As presented in Table~\ref{tab_CCD}, compared with (a) the original Chamfer Distance loss that employs the $L_2$-norm penalty, incorporating robust penalty functions, such as (b) the $L_1$-norm,  (c) the Welsch-Leclerc penalty, and (d) the Geman-McClure penalty, as the distance metric leads to a significant reduction in prediction errors across all speed groups.
In particular, (d) our Chamfer-GM loss using the Geman-McClure penalty surpasses (a) the original Chamfer Distance loss by 85\%, 51\%, and 27\% in the three speed groups, respectively.
Therefore, we adopt the Geman-McClure penalty as the default robust penalty function of our RCCD loss.
 
When adding point clouds from the past frame as part of the target data, (e) our multi-frame Chamfer-GM loss further decreases the error for the fast speed group from 1.66m to 1.46m. 
Moreover, building upon (e) the multi-frame loss, by using multi-frame consistency for reweighting, (g) our Robust Consistency-aware Chamfer loss, i.e., RCCD loss, decreases the error for  the static group from 0.08m to 0.06m. 
Particularly, compared to (f)~the CCD loss proposed in our original version, WeakMotion~\cite{li2023weakly},  (g)~our RCCD loss achieves a 40\% error reduction in the static group. 
Additionally, by integrating a smoothness regularization into the RCCD loss, (h) our final self-supervised motion loss function decreases the prediction errors of the three groups by 62\%, 39\%, and 27\%, respectively, compared to (f) the CCD loss.
The results demonstrate that our proposed RCCD loss and its smoothness regularization yield significant enhancements over the original CCD loss in WeakMotion~\cite{li2023weakly}.

Furthermore, to regularize the predicted motion, we also use an auxiliary FG/BG segmentation head for WeakMotionNet and set the motion of predicted background areas to zero.
As shown in Table~\ref{tab_CCD}, by (i) combining our  self-supervised motion loss with a FG/BG segmentation loss, we observe a significantly lower error on static group.

\begin{figure*}[t]
	\centering 
	\includegraphics[width=1\textwidth]{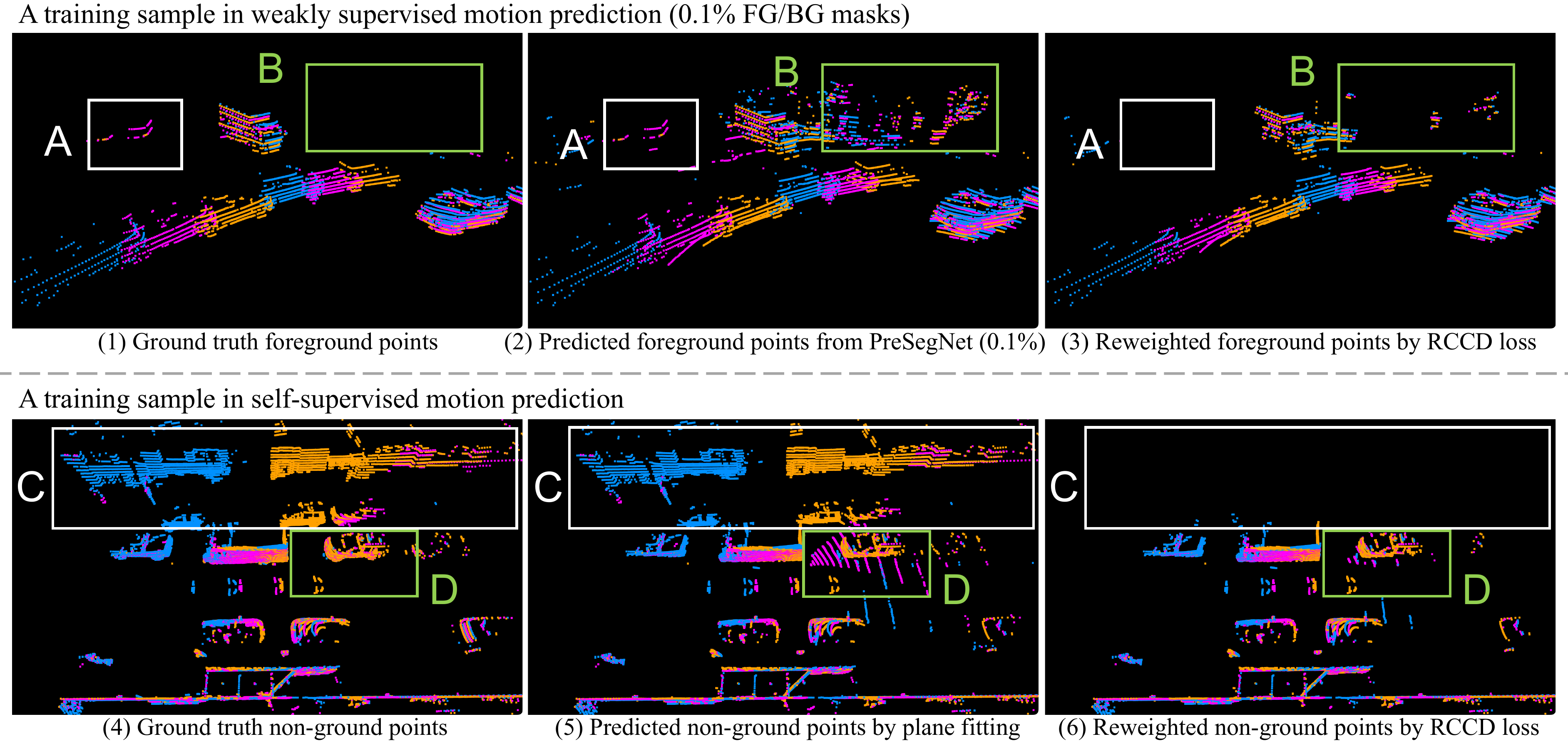}
	\vspace{-6mm}
	\caption{Visualization for PreSegNet, plane fitting, and RCCD loss. 
		Outliers may be due to occlusions of points (e.g., regions A and C),  inaccurate foreground predictions from PreSegNet (e.g., region B), and inaccurate non-ground predictions by the optimization-based plane fitting (e.g., region~D).
		To alleviate their impact, our RCCD loss uses multi-frame consistency to measure the confidence of points and assigns uncertain points less weight, thereby suppressing potential outliers. For better visualization, we remove points with lower weights in (3) and (6). Different color represents point cloud in different frames. 
		{\textcolor[RGB]{46,117,182}{\textbf{Blue}}}: past frame; 
		{\textcolor[RGB]{255,0,255}{\textbf{Purple}}}: current frame;
		{\textcolor[RGB]{237,125,49}{\textbf{Orange}}}: future frame.}
	\label{fig_chamfer}	
\end{figure*}

\subsubsection{Ablation study for weakly supervised  approaches.}
To enable weakly supervised motion learning using partially annotated FG/BG information, we develop two approaches: \textbf{WeakMotion-FB}, tailored for per-frame partially annotated FG/BG masks (1\% and 0.1\%), and  \textbf{WeakMotion-NG}, designed for partial FG/BG annotation of a single frame per sequence (0.01\%).

In \textbf{WeakMotion-FB}, the FG/BG segmentation network, PreSegNet, is trained with these incomplete FG/BG masks.
Subsequently, PreSegNet generates dense FG/BG masks for the self-supervised motion learning of WeakMotionNet. 
This allows us to derive dynamic motion supervision from the predicted foreground points.  
In \textbf{WeakMotion-NG}, we employ optimization-based plane fitting to predict the non-ground points for the self-supervised motion learning of WeakMotionNet. 
This enables us to derive dynamic motion supervision from the non-ground points, providing an alternative to foreground points.
Specifically, in the two approaches, an auxiliary FG/BG segmentation head is incorporated into WeakMotionNet to regularize motion prediction.

The ablation results of our two weakly supervised approaches are shown in Table~\ref{tab_Stage}.
Without predicting non-ground points or FG points, an alternative method is to treat every point as mobile and directly derive motion supervision from raw points in each frame. 
As demonstrated in Table~\ref{tab_Stage}, both WeakMotion-FB and WeakMotion-NG significantly outperform this alternative method across all evaluation metrics.
This indicates that  deriving motion supervision from predicted non-ground points and foreground points is more effective than directly using raw points.

In addition, the results in Table~\ref{tab_Stage} show that our \mbox{WeakMotion-FB} performs better than our WeakMotion-NG with 1\%  FG/BG annotations, while performing worse with 0.01\%  FG/BG annotations.
This discrepancy arises  because, in our \mbox{WeakMotion-FB}, the self-supervised motion learning relies on the FG/BG segmentation produced by the pretrained PreSegNet, which makes inaccurate FG/BG segmentation hinder the motion learning.
We evaluate the FG/BG segmentation performance of PreSegNet in \mbox{WeakMotion-FB}  on the nuScences validation set.
As shown in Table~\ref{tab_PreSegNet}, PreSegNet trained with 1\% FG/BG masks achieves a good overall segmentation accuracy of approximately 90\%, while the accuracy decreases to about 85\% when trained with 0.01\% FG/BG masks.
Unlike \mbox{WeakMotion-FB}, our \mbox{WeakMotion-NG} derives motion supervision from the predicted non-ground points, with FG/BG annotations used solely to train the auxiliary FG/BG segmentation head.
As a result, \mbox{WeakMotion-NG} requires fewer FG/BG annotations than \mbox{WeakMotion-FB} while  still delivering satisfactory performance.

\begin{table}
	\caption{ Foreground/background segmentation results produced by PreSegNet in WeakMotion-FB on the nuScenes validation set.}
	\vspace{-8mm}
	\label{tab_PreSegNet}
	\begin{center}
		\setlength{\tabcolsep}{3.0pt}
		\renewcommand\arraystretch{1.1}
		\resizebox{1\columnwidth}{!}{
			\centering
			\begin{tabular}{l@{\hskip 0.2cm}|c|c|c}
				\Xhline{1.2pt}
				Model & {BG  Acc.~$\uparrow$} &{FG Acc.~$\uparrow$}   & {Overall Acc.~$\uparrow$}  \\
				\Xhline{1.2pt}
				PreSegNet (1\% FG/BG masks)&  89.8\%   &  94.1\%   &  90.0\% \\
				PreSegNet (0.01\% FG/BG masks)&  85.6\% & 79.5\% & 85.3\%   \\
				\Xhline{1.2pt}
			\end{tabular}
		}
	\end{center}
\end{table}

\subsubsection{Ablation study for self-supervised approach.}
Without any annotations, we propose a self-supervised motion prediction approach, \textbf{SelfMotion-NG}.
Specifically, we predict non-ground points by an optimization-based plane fitting and derive motion supervision from the predicted non-ground points.
Without predicting non-ground points, an alternative method is to directly derive motion supervision from raw points. 
Table~\ref{tab_Self} presents a comparison between \mbox{SelfMotion-NG} and this alternative method.
As presented in Table~\ref{tab_Self}, our self-supervised approach, \mbox{SelfMotion-NG}, consistently outperforms this alternative method across all speed groups, demonstrating the effectiveness of our design.

\begin{table}
	\caption{Effectiveness of our self-supervised approach, SelfMotion-NG. Slow means speed~$\leq 5{\rm m/s}$. Fast means  speed~$\geq 5{\rm m/s}$. }
	\vspace{-8mm}
	\label{tab_Self}
	\begin{center}
		\setlength{\tabcolsep}{3.0pt}
		\renewcommand\arraystretch{1.1}
		\resizebox{1\columnwidth}{!}{
			\centering
			\begin{tabular}{l@{\hskip 0.05cm}|ccc}
				\Xhline{1.2pt}
				Self-supervised method & {Static$\downarrow$} &{Slow$\downarrow$}  & {Fast$\downarrow$}  \\
				\Xhline{1.2pt}
				Supervision from raw points & 1.721 	& 0.529 & 6.123    \\
				Supervision from non-ground points ({\bf SelfMotion-NG})  & \bf 0.044 & \bf 0.266  &\bf  2.184    \\
				\Xhline{1.2pt}
			\end{tabular}
		}
	\end{center}
\end{table}

\subsubsection{Visualization Analysis.}
In Fig.~\ref{fig_chamfer}, we provide two visualization examples of training data.
In our weakly and self-supervised motion prediction, outliers may be due to occlusion of points (e.g., regions A and C), inaccurate foreground predictions from PreSegNet (e.g., region~B), and inaccurate non-ground predictions by the optimization-based plane fitting (e.g., region~D), which may further impair the motion learning.
To address this issue, in our RCCD loss, we use multi-frame consistency to measure the confidence of points and assign uncertain points less weight, thereby suppressing potential outliers.  
As shown in Fig.~\ref{fig_chamfer}~(3) and~(6), the number of outliers in the four regions is reduced.

\section{Conclusion}
In this work, we study weakly and self-supervised class-agnostic motion prediction.
By associating motion understanding with scene parsing, we present a weakly supervised motion prediction approach, \mbox{WeakMotion-FB}, using fully or partially annotated FG/BG masks as weak supervision.
Additionally, by utilizing non-ground/ground masks as an alternative to FG/BG masks, we propose another weakly supervised approach, \mbox{WeakMotion-NG}, and a self-supervised approach, \mbox{SelfMotion-NG}, to further reduce or even eliminate FG/BG annotation effort.
We also design a novel Robust Consistency-aware Chamfer Distance loss for robust self-supervised motion learning.
Experiments show that our weakly supervised models surpass self-supervised ones and perform on par with some supervised ones, striking a good compromise between annotation effort and performance.
Moreover, without using any annotations for supervision, our self-supervised models achieve state-of-the-art performance in self-supervised motion prediction.

\section*{Acknowledgments}
This work was supported by the MOE AcRF Tier 2 grant (MOE-T2EP20223-0001) and the RIE2020 Industry Alignment Fund - Industry Collaboration Projects (IAF-ICP) Funding Initiative, as well as cash and in-kind contributions from industry partner(s). This work was also supported by the Jinan-NTU Green Technology Research Institute (GreenTRI), Nanyang Technological University, Singapore.

\bibliographystyle{ieeetr}
\bibliography{egbib}

\begin{thebibliography}{10}

\bibitem{menze2015object}
M.~Menze and A.~Geiger, ``Object scene flow for autonomous vehicles,'' in {\em
  Proceedings of the IEEE conference on computer vision and pattern
  recognition}, pp.~3061--3070, 2015.

\bibitem{fang2020tpnet}
L.~Fang, Q.~Jiang, J.~Shi, and B.~Zhou, ``Tpnet: Trajectory proposal network
  for motion prediction,'' in {\em Proceedings of the IEEE/CVF Conference on
  Computer Vision and Pattern Recognition}, pp.~6797--6806, 2020.

\bibitem{zhao2021tnt}
H.~Zhao, J.~Gao, T.~Lan, C.~Sun, B.~Sapp, B.~Varadarajan, Y.~Shen, Y.~Shen,
  Y.~Chai, C.~Schmid, {\em et~al.}, ``Tnt: Target-driven trajectory
  prediction,'' in {\em Conference on Robot Learning}, pp.~895--904, PMLR,
  2021.

\bibitem{gao2020vectornet}
J.~Gao, C.~Sun, H.~Zhao, Y.~Shen, D.~Anguelov, C.~Li, and C.~Schmid,
  ``Vectornet: Encoding hd maps and agent dynamics from vectorized
  representation,'' in {\em Proceedings of the IEEE/CVF Conference on Computer
  Vision and Pattern Recognition}, pp.~11525--11533, 2020.

\bibitem{wu2020motionnet}
P.~Wu, S.~Chen, and D.~N. Metaxas, ``Motionnet: Joint perception and motion
  prediction for autonomous driving based on bird's eye view maps,'' in {\em
  Proceedings of the IEEE/CVF conference on computer vision and pattern
  recognition}, pp.~11385--11395, 2020.

\bibitem{filatov2020any}
A.~Filatov, A.~Rykov, and V.~Murashkin, ``Any motion detector: Learning
  class-agnostic scene dynamics from a sequence of lidar point clouds,'' in
  {\em 2020 IEEE international conference on robotics and automation (ICRA)},
  pp.~9498--9504, IEEE, 2020.

\bibitem{wang2022sti}
Y.~Wang, H.~Pan, J.~Zhu, Y.-H. Wu, X.~Zhan, K.~Jiang, and D.~Yang, ``Be-sti:
  Spatial-temporal integrated network for class-agnostic motion prediction with
  bidirectional enhancement,'' in {\em Proceedings of the IEEE/CVF Conference
  on Computer Vision and Pattern Recognition}, pp.~17093--17102, 2022.

\bibitem{geiger2012we}
A.~Geiger, P.~Lenz, and R.~Urtasun, ``Are we ready for autonomous driving? the
  kitti vision benchmark suite,'' in {\em 2012 IEEE conference on computer
  vision and pattern recognition}, pp.~3354--3361, IEEE, 2012.

\bibitem{jund2021scalable}
P.~Jund, C.~Sweeney, N.~Abdo, Z.~Chen, and J.~Shlens, ``Scalable scene flow
  from point clouds in the real world,'' {\em IEEE Robotics and Automation
  Letters}, vol.~7, no.~2, pp.~1589--1596, 2021.

\bibitem{wang2024semi}
K.~Wang, Y.~Wu, Z.~Pan, X.~Li, K.~Xian, Z.~Wang, Z.~Cao, and G.~Lin,
  ``Semi-supervised class-agnostic motion prediction with pseudo label
  regeneration and bevmix,'' in {\em Proceedings of the AAAI Conference on
  Artificial Intelligence}, vol.~38, pp.~5490--5498, 2024.

\bibitem{luo2021self}
C.~Luo, X.~Yang, and A.~Yuille, ``Self-supervised pillar motion learning for
  autonomous driving,'' in {\em Proceedings of the IEEE/CVF Conference on
  Computer Vision and Pattern Recognition}, pp.~3183--3192, 2021.

\bibitem{jia2023contrastmotion}
X.~Jia, H.~Zhou, X.~Zhu, Y.~Guo, J.~Zhang, and Y.~Ma, ``Contrastmotion:
  Self-supervised scene motion learning for large-scale lidar point clouds,''
  in {\em Proceedings of the Thirty-Second International Joint Conference on
  Artificial Intelligence}, pp.~929--937, 2023.

\bibitem{wang2024self}
K.~Wang, Y.~Wu, J.~Cen, Z.~Pan, X.~Li, Z.~Wang, Z.~Cao, and G.~Lin,
  ``Self-supervised class-agnostic motion prediction with spatial and temporal
  consistency regularizations,'' in {\em Proceedings of the IEEE/CVF Conference
  on Computer Vision and Pattern Recognition (CVPR)}, pp.~14638--14647, June
  2024.

\bibitem{fang2024self}
S.~Fang, Z.~Liu, M.~Wang, C.~Xu, Y.~Zhong, and S.~Chen, ``Self-supervised
  bird’s eye view motion prediction with cross-modality signals,'' in {\em
  Proceedings of the AAAI Conference on Artificial Intelligence}, vol.~38,
  pp.~1726--1734, 2024.

\bibitem{wu2020pointpwc}
W.~Wu, Z.~Y. Wang, Z.~Li, W.~Liu, and L.~Fuxin, ``Pointpwc-net: Cost volume on
  point clouds for (self-) supervised scene flow estimation,'' in {\em European
  conference on computer vision}, pp.~88--107, Springer, 2020.

\bibitem{kittenplon2021flowstep3d}
Y.~Kittenplon, Y.~C. Eldar, and D.~Raviv, ``Flowstep3d: Model unrolling for
  self-supervised scene flow estimation,'' in {\em Proceedings of the IEEE/CVF
  Conference on Computer Vision and Pattern Recognition}, pp.~4114--4123, 2021.

\bibitem{tatarchenko2019single}
M.~Tatarchenko, S.~R. Richter, R.~Ranftl, Z.~Li, V.~Koltun, and T.~Brox, ``What
  do single-view 3d reconstruction networks learn?,'' in {\em Proceedings of
  the IEEE/CVF conference on computer vision and pattern recognition},
  pp.~3405--3414, 2019.

\bibitem{geman1985}
S.~Geman and D.~E. McClure, ``Bayesian image analysis: An application to single
  photon emission tomography,'' in {\em Proceedings of the American Statistical
  Association}, 1985.

\bibitem{barron2019general}
J.~T. Barron, ``A general and adaptive robust loss function,'' in {\em
  Proceedings of the IEEE/CVF conference on computer vision and pattern
  recognition}, pp.~4331--4339, 2019.

\bibitem{li2023weakly}
R.~Li, H.~Shi, Z.~Fu, Z.~Wang, and G.~Lin, ``Weakly supervised class-agnostic
  motion prediction for autonomous driving,'' in {\em Proceedings of the
  IEEE/CVF Conference on Computer Vision and Pattern Recognition},
  pp.~17599--17608, 2023.

\bibitem{ettinger2021large}
S.~Ettinger, S.~Cheng, B.~Caine, C.~Liu, H.~Zhao, S.~Pradhan, Y.~Chai, B.~Sapp,
  C.~R. Qi, Y.~Zhou, {\em et~al.}, ``Large scale interactive motion forecasting
  for autonomous driving: The waymo open motion dataset,'' in {\em Proceedings
  of the IEEE/CVF International Conference on Computer Vision}, pp.~9710--9719,
  2021.

\bibitem{phan2020covernet}
T.~Phan-Minh, E.~C. Grigore, F.~A. Boulton, O.~Beijbom, and E.~M. Wolff,
  ``Covernet: Multimodal behavior prediction using trajectory sets,'' in {\em
  Proceedings of the IEEE/CVF Conference on Computer Vision and Pattern
  Recognition}, pp.~14074--14083, 2020.

\bibitem{gu2021densetnt}
J.~Gu, C.~Sun, and H.~Zhao, ``Densetnt: End-to-end trajectory prediction from
  dense goal sets,'' in {\em Proceedings of the IEEE/CVF International
  Conference on Computer Vision}, pp.~15303--15312, 2021.

\bibitem{zhang2020stinet}
Z.~Zhang, J.~Gao, J.~Mao, Y.~Liu, D.~Anguelov, and C.~Li, ``Stinet:
  Spatio-temporal-interactive network for pedestrian detection and trajectory
  prediction,'' in {\em Proceedings of the IEEE/CVF Conference on Computer
  Vision and Pattern Recognition}, pp.~11346--11355, 2020.

\bibitem{liang2020pnpnet}
M.~Liang, B.~Yang, W.~Zeng, Y.~Chen, R.~Hu, S.~Casas, and R.~Urtasun, ``Pnpnet:
  End-to-end perception and prediction with tracking in the loop,'' in {\em
  Proceedings of the IEEE/CVF Conference on Computer Vision and Pattern
  Recognition}, pp.~11553--11562, 2020.

\bibitem{chang2019argoverse}
M.-F. Chang, J.~Lambert, P.~Sangkloy, J.~Singh, S.~Bak, A.~Hartnett, D.~Wang,
  P.~Carr, S.~Lucey, D.~Ramanan, {\em et~al.}, ``Argoverse: 3d tracking and
  forecasting with rich maps,'' in {\em Proceedings of the IEEE/CVF Conference
  on Computer Vision and Pattern Recognition}, pp.~8748--8757, 2019.

\bibitem{lee2020pillarflow}
K.-H. Lee, M.~Kliemann, A.~Gaidon, J.~Li, C.~Fang, S.~Pillai, and W.~Burgard,
  ``Pillarflow: End-to-end birds-eye-view flow estimation for autonomous
  driving,'' in {\em 2020 IEEE/RSJ International Conference on Intelligent
  Robots and Systems (IROS)}, pp.~2007--2013, IEEE, 2020.

\bibitem{schreiber2021dynamic}
M.~Schreiber, V.~Belagiannis, C.~Gl{\"a}ser, and K.~Dietmayer, ``Dynamic
  occupancy grid mapping with recurrent neural networks,'' in {\em 2021 IEEE
  International Conference on Robotics and Automation (ICRA)}, pp.~6717--6724,
  IEEE, 2021.

\bibitem{sun2018pwc}
D.~Sun, X.~Yang, M.-Y. Liu, and J.~Kautz, ``Pwc-net: Cnns for optical flow
  using pyramid, warping, and cost volume,'' in {\em Proceedings of the IEEE
  conference on computer vision and pattern recognition}, pp.~8934--8943, 2018.

\bibitem{li2024self}
R.~Li, C.~Zhang, Z.~Wang, C.~Shen, and G.~Lin, ``Self-supervised 3d scene flow
  estimation and motion prediction using local rigidity prior,'' {\em IEEE
  Transactions on Pattern Analysis and Machine Intelligence}, 2024.

\bibitem{vedula1999three}
S.~Vedula, S.~Baker, P.~Rander, R.~Collins, and T.~Kanade, ``Three-dimensional
  scene flow,'' in {\em Proceedings of the Seventh IEEE International
  Conference on Computer Vision}, vol.~2, pp.~722--729, IEEE, 1999.

\bibitem{liu2019flownet3d}
X.~Liu, C.~R. Qi, and L.~J. Guibas, ``Flownet3d: Learning scene flow in 3d
  point clouds,'' in {\em Proceedings of the IEEE/CVF conference on computer
  vision and pattern recognition}, pp.~529--537, 2019.

\bibitem{cheng2022bi}
W.~Cheng and J.~H. Ko, ``Bi-pointflownet: Bidirectional learning for point
  cloud based scene flow estimation,'' in {\em European Conference on Computer
  Vision}, pp.~108--124, Springer, 2022.

\bibitem{li2021hcrf}
R.~Li, G.~Lin, T.~He, F.~Liu, and C.~Shen, ``Hcrf-flow: Scene flow from point
  clouds with continuous high-order crfs and position-aware flow embedding,''
  in {\em Proceedings of the IEEE/CVF Conference on Computer Vision and Pattern
  Recognition}, pp.~364--373, 2021.

\bibitem{li2022rigidflow}
R.~Li, C.~Zhang, G.~Lin, Z.~Wang, and C.~Shen, ``Rigidflow: Self-supervised
  scene flow learning on point clouds by local rigidity prior,'' in {\em
  Proceedings of the IEEE/CVF Conference on Computer Vision and Pattern
  Recognition}, pp.~16959--16968, 2022.

\bibitem{li2021self}
R.~Li, G.~Lin, and L.~Xie, ``Self-point-flow: Self-supervised scene flow
  estimation from point clouds with optimal transport and random walk,'' in
  {\em Proceedings of the IEEE/CVF conference on computer vision and pattern
  recognition}, pp.~15577--15586, 2021.

\bibitem{gojcic2021weakly}
Z.~Gojcic, O.~Litany, A.~Wieser, L.~J. Guibas, and T.~Birdal, ``Weakly
  supervised learning of rigid 3d scene flow,'' in {\em Proceedings of the
  IEEE/CVF conference on computer vision and pattern recognition},
  pp.~5692--5703, 2021.

\bibitem{dong2022exploiting}
G.~Dong, Y.~Zhang, H.~Li, X.~Sun, and Z.~Xiong, ``Exploiting rigidity
  constraints for lidar scene flow estimation,'' in {\em Proceedings of the
  IEEE/CVF Conference on Computer Vision and Pattern Recognition},
  pp.~12776--12785, 2022.

\bibitem{wang2022matters}
G.~Wang, Y.~Hu, Z.~Liu, Y.~Zhou, M.~Tomizuka, W.~Zhan, and H.~Wang, ``What
  matters for 3d scene flow network,'' in {\em Computer Vision--ECCV 2022: 17th
  European Conference, Tel Aviv, Israel, October 23--27, 2022, Proceedings,
  Part XXXIII}, pp.~38--55, Springer, 2022.

\bibitem{mittal2020just}
H.~Mittal, B.~Okorn, and D.~Held, ``Just go with the flow: Self-supervised
  scene flow estimation,'' in {\em Proceedings of the IEEE/CVF Conference on
  Computer Vision and Pattern Recognition}, pp.~11177--11185, 2020.

\bibitem{tishchenko2020self}
I.~Tishchenko, S.~Lombardi, M.~R. Oswald, and M.~Pollefeys, ``Self-supervised
  learning of non-rigid residual flow and ego-motion,'' in {\em 2020
  International Conference on 3D Vision (3DV)}, pp.~150--159, IEEE, 2020.

\bibitem{baur2021slim}
S.~A. Baur, D.~J. Emmerichs, F.~Moosmann, P.~Pinggera, B.~Ommer, and A.~Geiger,
  ``Slim: Self-supervised lidar scene flow and motion segmentation,'' in {\em
  Proceedings of the IEEE/CVF International Conference on Computer Vision},
  pp.~13126--13136, 2021.

\bibitem{pontes2020scene}
J.~K. Pontes, J.~Hays, and S.~Lucey, ``Scene flow from point clouds with or
  without learning,'' in {\em 2020 international conference on 3D vision
  (3DV)}, pp.~261--270, IEEE, 2020.

\bibitem{gu2022rcp}
X.~Gu, C.~Tang, W.~Yuan, Z.~Dai, S.~Zhu, and P.~Tan, ``Rcp: Recurrent closest
  point for point cloud,'' in {\em Proceedings of the IEEE/CVF Conference on
  Computer Vision and Pattern Recognition}, pp.~8216--8226, 2022.

\bibitem{he2022self}
P.~He, P.~Emami, S.~Ranka, and A.~Rangarajan, ``Self-supervised robust scene
  flow estimation via the alignment of probability density functions,'' in {\em
  Proceedings of the AAAI Conference on Artificial Intelligence}, vol.~36,
  pp.~861--869, 2022.

\bibitem{shen2023self}
Y.~Shen, L.~Hui, J.~Xie, and J.~Yang, ``Self-supervised 3d scene flow
  estimation guided by superpoints,'' in {\em Proceedings of the IEEE/CVF
  Conference on Computer Vision and Pattern Recognition}, pp.~5271--5280, 2023.

\bibitem{li2021neural}
X.~Li, J.~Kaesemodel~Pontes, and S.~Lucey, ``Neural scene flow prior,'' {\em
  Advances in Neural Information Processing Systems}, vol.~34, pp.~7838--7851,
  2021.

\bibitem{xiao2023unsupervised}
A.~Xiao, J.~Huang, D.~Guan, X.~Zhang, S.~Lu, and L.~Shao, ``Unsupervised point
  cloud representation learning with deep neural networks: A survey,'' {\em
  IEEE Transactions on Pattern Analysis and Machine Intelligence}, vol.~45,
  no.~9, pp.~11321--11339, 2023.

\bibitem{xie2020pointcontrast}
S.~Xie, J.~Gu, D.~Guo, C.~R. Qi, L.~Guibas, and O.~Litany, ``Pointcontrast:
  Unsupervised pre-training for 3d point cloud understanding,'' in {\em
  Computer Vision--ECCV 2020: 16th European Conference, Glasgow, UK, August
  23--28, 2020, Proceedings, Part III 16}, pp.~574--591, Springer, 2020.

\bibitem{sanghi2020info3d}
A.~Sanghi, ``Info3d: Representation learning on 3d objects using mutual
  information maximization and contrastive learning,'' in {\em Computer
  Vision--ECCV 2020: 16th European Conference, Glasgow, UK, August 23--28,
  2020, Proceedings, Part XXIX 16}, pp.~626--642, Springer, 2020.

\bibitem{zhang2021self}
Z.~Zhang, R.~Girdhar, A.~Joulin, and I.~Misra, ``Self-supervised pretraining of
  3d features on any point-cloud,'' in {\em Proceedings of the IEEE/CVF
  international conference on computer vision}, pp.~10252--10263, 2021.

\bibitem{yin2022proposalcontrast}
J.~Yin, D.~Zhou, L.~Zhang, J.~Fang, C.-Z. Xu, J.~Shen, and W.~Wang,
  ``Proposalcontrast: Unsupervised pre-training for lidar-based 3d object
  detection,'' in {\em European conference on computer vision}, pp.~17--33,
  Springer, 2022.

\bibitem{liu2022masked}
H.~Liu, M.~Cai, and Y.~J. Lee, ``Masked discrimination for self-supervised
  learning on point clouds,'' in {\em European Conference on Computer Vision},
  pp.~657--675, Springer, 2022.

\bibitem{yu2022point}
X.~Yu, L.~Tang, Y.~Rao, T.~Huang, J.~Zhou, and J.~Lu, ``Point-bert:
  Pre-training 3d point cloud transformers with masked point modeling,'' in
  {\em Proceedings of the IEEE/CVF conference on computer vision and pattern
  recognition}, pp.~19313--19322, 2022.

\bibitem{pang2022masked}
Y.~Pang, W.~Wang, F.~E. Tay, W.~Liu, Y.~Tian, and L.~Yuan, ``Masked
  autoencoders for point cloud self-supervised learning,'' in {\em European
  conference on computer vision}, pp.~604--621, Springer, 2022.

\bibitem{wang2021unsupervised}
H.~Wang, Q.~Liu, X.~Yue, J.~Lasenby, and M.~J. Kusner, ``Unsupervised point
  cloud pre-training via occlusion completion,'' in {\em Proceedings of the
  IEEE/CVF international conference on computer vision}, pp.~9782--9792, 2021.

\bibitem{zhang2022point}
R.~Zhang, Z.~Guo, P.~Gao, R.~Fang, B.~Zhao, D.~Wang, Y.~Qiao, and H.~Li,
  ``Point-m2ae: multi-scale masked autoencoders for hierarchical point cloud
  pre-training,'' {\em Advances in neural information processing systems},
  vol.~35, pp.~27061--27074, 2022.

\bibitem{huang2023ponder}
D.~Huang, S.~Peng, T.~He, H.~Yang, X.~Zhou, and W.~Ouyang, ``Ponder: Point
  cloud pre-training via neural rendering,'' in {\em Proceedings of the
  IEEE/CVF International Conference on Computer Vision}, pp.~16089--16098,
  2023.

\bibitem{chen2023pointgpt}
G.~Chen, M.~Wang, Y.~Yang, K.~Yu, L.~Yuan, and Y.~Yue, ``Pointgpt:
  Auto-regressively generative pre-training from point clouds,'' {\em Advances
  in Neural Information Processing Systems}, vol.~36, pp.~29667--29679, 2023.

\bibitem{wu2021density}
T.~Wu, L.~Pan, J.~Zhang, T.~Wang, Z.~Liu, and D.~Lin, ``Density-aware chamfer
  distance as a comprehensive metric for point cloud completion,'' {\em arXiv
  preprint arXiv:2111.12702}, 2021.

\bibitem{yang2019pointflow}
G.~Yang, X.~Huang, Z.~Hao, M.-Y. Liu, S.~Belongie, and B.~Hariharan,
  ``Pointflow: 3d point cloud generation with continuous normalizing flows,''
  in {\em Proceedings of the IEEE/CVF International Conference on Computer
  Vision}, pp.~4541--4550, 2019.

\bibitem{choe2021deep}
J.~Choe, B.~Joung, F.~Rameau, J.~Park, and I.~S. Kweon, ``Deep point cloud
  reconstruction,'' {\em arXiv preprint arXiv:2111.11704}, 2021.

\bibitem{wang2022neural}
C.~Wang, X.~Li, J.~K. Pontes, and S.~Lucey, ``Neural prior for trajectory
  estimation,'' in {\em Proceedings of the IEEE/CVF Conference on Computer
  Vision and Pattern Recognition}, pp.~6532--6542, 2022.

\bibitem{ouyang2021occlusion}
B.~Ouyang and D.~Raviv, ``Occlusion guided self-supervised scene flow
  estimation on 3d point clouds,'' in {\em 2021 International Conference on 3D
  Vision (3DV)}, pp.~782--791, IEEE, 2021.

\bibitem{he2022learning}
P.~He, P.~Emami, S.~Ranka, and A.~Rangarajan, ``Learning scene dynamics from
  point cloud sequences,'' {\em International Journal of Computer Vision},
  vol.~130, no.~3, pp.~669--695, 2022.

\bibitem{dennis1978techniques}
J.~E. Dennis~Jr and R.~E. Welsch, ``Techniques for nonlinear least squares and
  robust regression,'' {\em Communications in Statistics-simulation and
  Computation}, vol.~7, no.~4, pp.~345--359, 1978.

\bibitem{leclerc1989constructing}
Y.~G. Leclerc, ``Constructing simple stable descriptions for image
  partitioning,'' {\em International journal of computer vision}, vol.~3,
  no.~1, pp.~73--102, 1989.

\bibitem{najibi2022motion}
M.~Najibi, J.~Ji, Y.~Zhou, C.~R. Qi, X.~Yan, S.~Ettinger, and D.~Anguelov,
  ``Motion inspired unsupervised perception and prediction in autonomous
  driving,'' in {\em European Conference on Computer Vision}, pp.~424--443,
  Springer, 2022.

\bibitem{chodosh2023re}
N.~Chodosh, D.~Ramanan, and S.~Lucey, ``Re-evaluating lidar scene flow for
  autonomous driving,'' {\em arXiv preprint arXiv:2304.02150}, 2023.

\bibitem{fischler1981random}
M.~A. Fischler and R.~C. Bolles, ``Random sample consensus: a paradigm for
  model fitting with applications to image analysis and automated
  cartography,'' {\em Communications of the ACM}, vol.~24, no.~6, pp.~381--395,
  1981.

\bibitem{caesar2020nuscenes}
H.~Caesar, V.~Bankiti, A.~H. Lang, S.~Vora, V.~E. Liong, Q.~Xu, A.~Krishnan,
  Y.~Pan, G.~Baldan, and O.~Beijbom, ``nuscenes: A multimodal dataset for
  autonomous driving,'' in {\em Proceedings of the IEEE/CVF conference on
  computer vision and pattern recognition}, pp.~11621--11631, 2020.

\bibitem{sun2020scalability}
P.~Sun, H.~Kretzschmar, X.~Dotiwalla, A.~Chouard, V.~Patnaik, P.~Tsui, J.~Guo,
  Y.~Zhou, Y.~Chai, B.~Caine, {\em et~al.}, ``Scalability in perception for
  autonomous driving: Waymo open dataset,'' in {\em Proceedings of the IEEE/CVF
  conference on computer vision and pattern recognition}, pp.~2446--2454, 2020.

\bibitem{gu2019hplflownet}
X.~Gu, Y.~Wang, C.~Wu, Y.~J. Lee, and P.~Wang, ``Hplflownet: Hierarchical
  permutohedral lattice flownet for scene flow estimation on large-scale point
  clouds,'' in {\em Proceedings of the IEEE/CVF conference on computer vision
  and pattern recognition}, pp.~3254--3263, 2019.

\bibitem{shi2019pointrcnn}
S.~Shi, X.~Wang, and H.~Li, ``Pointrcnn: 3d object proposal generation and
  detection from point cloud,'' in {\em Proceedings of the IEEE/CVF conference
  on computer vision and pattern recognition}, pp.~770--779, 2019.

\bibitem{schreiber2019long}
M.~Schreiber, S.~Hoermann, and K.~Dietmayer, ``Long-term occupancy grid
  prediction using recurrent neural networks,'' in {\em 2019 International
  Conference on Robotics and Automation (ICRA)}, pp.~9299--9305, IEEE, 2019.

\bibitem{kingma2014adam}
D.~P. Kingma and J.~Ba, ``Adam: A method for stochastic optimization,'' {\em
  arXiv preprint arXiv:1412.6980}, 2014.

\bibitem{paszke2019pytorch}
A.~Paszke, S.~Gross, F.~Massa, A.~Lerer, J.~Bradbury, G.~Chanan, T.~Killeen,
  Z.~Lin, N.~Gimelshein, L.~Antiga, {\em et~al.}, ``Pytorch: An imperative
  style, high-performance deep learning library,'' {\em Advances in neural
  information processing systems}, vol.~32, 2019.

\end{thebibliography}

\end{document}